%% file: graph_da.tex
\documentclass{elektr}
\usepackage{hyperref}
\hypersetup{
colorlinks=true,
urlcolor=blue,
citecolor=blue}
\usepackage[all]{xy,xypic}
\usepackage{amsfonts,amssymb,amsmath,amsgen,amsopn,amsbsy,theorem,graphicx,epsfig,subfigure}
\usepackage{eufrak,amscd,bezier,latexsym,mathrsfs,enumerate}\usepackage[utf8]{inputenc}\usepackage[english]{babel}
\usepackage[dvipsnames]{xcolor}
\usepackage[pagewise]{lineno}
\usepackage{algorithm}
\usepackage{algorithmic}
\usepackage{float}

\year{}
\vol{}
\fpage{}
\lpage{}
\doi{}

\title{Domain Adaptation on Graphs by Learning Graph Topologies: Theoretical Analysis and an Algorithm}

\author[Elif Vural]{
\textbf{Elif Vural$^{1}$\thanks{velif@metu.edu.tr. This work has been supported by the T\"UB\.ITAK 2232 program, project no.~117C007.
}~}\\
$^{1}$Department of Electrical and Electronics Engineering, Middle East Technical University, Ankara, Turkey\\
\\ [1.8em]

}

\input{elksty.tex}

\begin{document}

\maketitle

\begin{abstract}

Traditional machine learning algorithms assume that the training and test data have the same distribution, while this assumption does not necessarily hold in real applications. Domain adaptation methods take into account the deviations in the data distribution. In this work, we study the problem of domain adaptation on graphs. We consider a source graph and a target graph constructed with samples drawn from data manifolds. We study the problem of estimating the unknown class labels on the target graph using the label information on the source graph and the similarity between the two graphs. We particularly focus on a setting where the target label function is learnt such that its spectrum  is similar to that of the source label function.  We first propose a theoretical analysis of domain adaptation on graphs and present performance bounds that characterize the target classification error in terms of the properties of the graphs and the data manifolds. We show that the classification performance improves as the topologies of the graphs get more balanced, i.e., as the numbers of neighbors of different graph nodes become more proportionate, and weak edges with small weights are avoided. Our results also suggest that graph edges between too distant data samples should be avoided for good generalization performance. We then propose a graph domain adaptation algorithm inspired by our theoretical findings, which estimates the label functions while learning the source and target graph topologies at the same time. The joint graph learning and label estimation problem is formulated through an objective function relying on our performance bounds, which is minimized with an alternating optimization scheme. Experiments on synthetic and real data sets suggest that the proposed method outperforms baseline approaches. 


\end{abstract}

\section{Introduction}
\label{Int}

Classical machine learning methods are based on the assumption that the training  data and the test data have the same distribution. A classifier is learnt on the training data, which is then used to estimate the unknown class labels of the test data. On the other hand, domain adaptation methods consider a setting where the distribution of the test data is different from that of the training data \cite{RohrbachES13, WangM08, XiaoG15, FangZ13}. Given many labeled samples in a source domain and much fewer labeled samples in a target domain, domain adaptation algorithms exploit the information available in both domains in order to improve the performance of classification in the target domain. Meanwhile, many machine learning applications nowadays involve inference problems on graph domains such as social networks and communication networks. Moreover, in most problems data samples conform to a low-dimensional manifold model; for instance, face images of a person captured from different camera angles lie on a low-dimensional manifold. In such problems, graphs models are widely used as they are very convenient for approximating the actual data manifolds. Hence, domain adaptation on graphs arises as an important problem of interest, which we study in this work. We first present performance bounds for transferring the knowledge of class labels between a pair of graphs. We then use these bounds to develop an algorithm that computes the structures of the source and the target graphs and estimates the unknown class labels at the same time.

\begin{figure}[t]
\begin{center}
     \subfigure[]
       {\label{fig:gen_face_manif}\includegraphics[width=3.5cm]{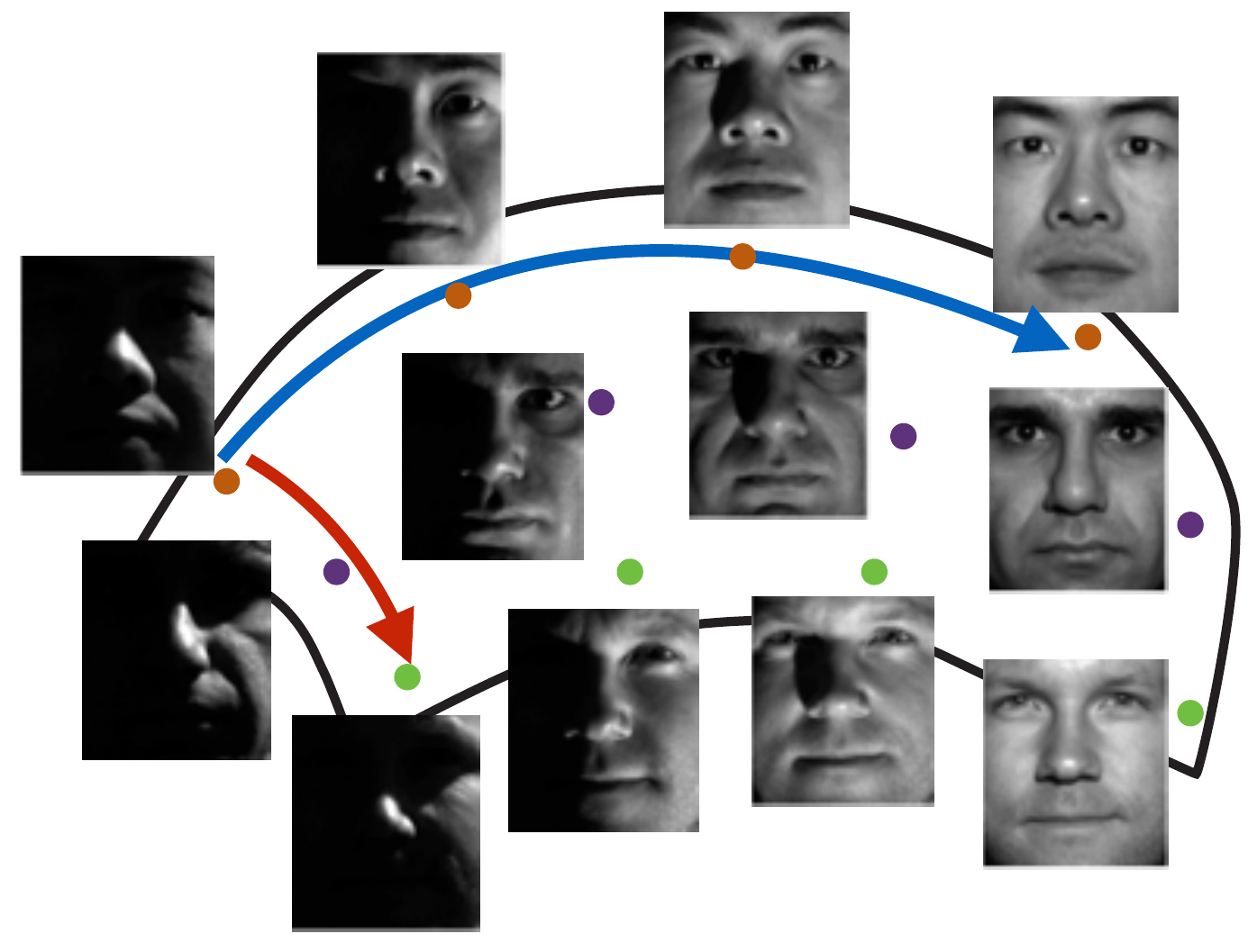}}
       \hspace{2cm}
      \subfigure[]
       {\label{fig:illus_da_graphs}\includegraphics[width=8cm]{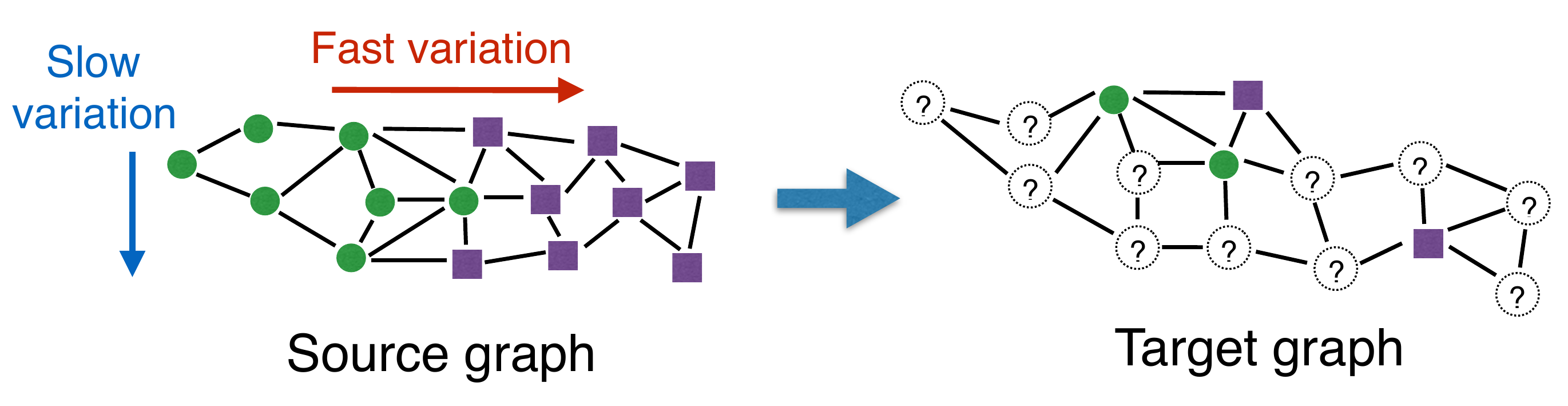}} 
 \end{center}
 \vspace{-0.5cm}
 \caption{(a) Illustration of a face manifold where the label function varies at different speeds along different directions  (b) Illustration of domain adaptation on graphs}
 \vspace{-0.5cm}
\end{figure}

Many graph-based learning methods rely on the assumption that the label function varies slowly on the data graph. However, this assumption does not always hold as shown in Figure \ref{fig:gen_face_manif}. A face manifold is illustrated in Figure \ref{fig:gen_face_manif}, each point of which corresponds to a face image that belongs to one of three different subjects. In this example, the class label function varies slowly along the blue direction, where images belong to the same subject. On the other hand, along the red direction the face images of different subjects get too close to each other due to extreme illumination conditions. Hence, the label function has fast variation along the red direction on the manifold.

Although it is common to assume that the label function varies slowly in problems concerning a single graph, in a problem with more than one graph it is possible to learn the speed of variation of the label function and share this information across different graphs. This is illustrated in Figure \ref{fig:illus_da_graphs}, where the characteristics of  the variation of the label function can be learnt on a source graph where many class labels are available. Then, the purpose of graph domain adaptation is to transfer this information to a target graph that contains very few labels and estimate the unknown class labels. We studied this problem in our previous work \cite{PilanciV16}, where we proposed a method called Spectral Domain Adaptation (SDA). 

In order to optimize the performance of graph domain adaptation methods, it is important to theoretically characterize their performance limits. The classification performance significantly depends on the structures of the source and the target graphs and the similarity between them. In particular, in problems where the graphs are constructed from data samples coming from data manifolds, the properties of the graphs such as the locations and the weights of the edges, and the number of neighbors of graph nodes largely influence the performance of learning. A thorough characterization of the effects of such parameters in conjunction with the geometry of the data manifolds is necessary to understand the performance limits of graph domain adaptation.

Our contribution in this study is twofold. We first propose a theoretical study of domain adaptation on graphs. We consider a source graph and a target graph constructed with data samples coming respectively from a source manifold and a target manifold. We theoretically analyze the performance of classification on the target graph. In particular, we focus on the estimation error of the target label function and analyze how this error varies with the graph properties, the sampling density of data, and the geometric properties of the data manifolds. Our theoretical analysis suggests that very sparse graphs with too few edges and very dense graphs with too many edges should be avoided, as well as too small edge weights. The smoothness of the label function is shown to positively influence the performance of learning. We show that, under certain assumptions, the estimation error of the target label function decreases with the sampling density of the manifolds at a rate of $O(\N^{-1/d})$, where $\N$ is the number of samples and $d$ is the intrinsic dimension of the manifolds. Next, we use these theoretical findings to propose a graph domain adaptation algorithm that jointly estimates the class labels of the source and the target data and the topologies of the source and the target graphs. In particular, we optimize the source and the target graph weight matrices, which fully determine the graph topologies, in order to properly control parameters such as the number of neighbors,  the minimum edge weights, and the smoothness of the label functions on the graphs. Experimental results on synthetic and real data sets show that the proposed method with learnt graph topologies outperforms reference domain adaptation methods with fixed graph topologies and other baseline algorithms. 
 
The rest of the paper is organized as follows. In Section \ref{sec:rel_work}, we briefly overview the related literature. In Section \ref{sec:theo_analy}, we first overview frequency analysis on graphs and then present theoretical bounds for graph domain adaptation. In Section \ref{sec:prop_method}, we present a graph domain adaptation algorithm that is motivated by our theoretical findings. We  experimentally evaluate the proposed method in Section \ref{sec:exp_res} and conclude in Section \ref{sec:concl}.

\section{Related Work}
\label{sec:rel_work}

We first overview some common approaches in domain adaptation. In several previous works the covariate shift problem is studied, where the two distributions are matched with reweighting  \cite{HuangSGBS06, KhalighiRN17}. The studies in \cite{Daume2010, DuanXT12} propose to train a common classifier after mapping the data to a higher dimensional domain via feature augmentation. Another common approach is to align or match the two domains by mapping them to a common domain with projections or transformations \cite{WangM11, GongSSG12, Fernando2013, GongZLTGS16, PanTKY11, CourtyFTR17, JiangHHY17, YanDLWXZ17}.

Several domain adaptation methods model data with a graph and make use of the assumption that the label function varies smoothly on the graph \cite{XiaoG15, YaoPNLM15, ChengP14}. The unsupervised method in \cite{DasL18} formulates the domain adaptation problem as a graph matching problem. The algorithms in \cite{EynardKBGB15} and \cite{RodolaCBTC17} aim to compute a pair of bases on the source graph and the target graph that approximate the Fourier bases and jointly diagonalize the two graph Laplacians. These methods are applied to problems such as clustering and 3D shape analysis. Our recent work \cite{PilanciV16} also relies on representations with graph Fourier bases, however, in the context of domain adaptation. Our study in this paper provides insights for such graph-based methods involving the notion of smoothness on graphs and representations in graph bases.

The problem of learning graph topologies from data has drawn particular interest in the recent years. Various efforts have focused on the inference of the graph topology from a set of training signals that are known to vary smoothly on the graph \cite{ DongTFV16, Kalofolias16, EgilmezPO17}. However, such approaches differ from ours in that they address an unsupervised learning problem. Our method, on the other hand, actively incorporates the information of the class labels when learning the graph structures in a domain adaptation framework. In some earlier studies, graph structures have been learnt in a semi-supervised setting \cite{ArgyriouHP05, DaiY07}. However,  unlike as in our work, the graph Laplacians are restricted to a linear combination of a prescribed set of kernels in these methods.

Some previous studies analyzing the domain adaptation problem from a theoretical perspective are the following. Performance bounds for importance reweighting have been proposed in \cite{HuangSGBS06}  and \cite{CortesMM10}. The studies in \cite{BenDavidBCP06, BlitzerCKPW07, MansourMR09, BenDavidBCKPV10, ZhangZY13} bound the target loss in terms of the deviation between the source distribution and the target distribution. While such studies present a theoretical analysis of domain adaptation, none of them treat the domain adaptation problem in a graph setting. To the best of our knowledge, our theoretical analysis is the first to focus particularly on the graph domain adaptation problem.

\section{Theoretical Analysis of Graph Domain Adaptation}
\label{sec:theo_analy}

\subsection{Overview of Signal Processing on Graphs}
\label{ssec:sgt}

We first briefly overview some basic concepts regarding spectral graph theory and signal processing on graphs \cite{Chung97, ShumanNFOV13}. Let $G=(V, E)$ be a graph consisting of $N$ vertices denoted by $V=\{ x_i \}_{i=1}^\N$ and edges $E$. The $\N \times \N$ symmetric matrix $W$ consisting of nonnegative edge weights is called the weight matrix, where $W_{ij} $ is the weight of the edge between the nodes $x_i$ and $x_j$. If there is no edge between  $x_i$ and $x_j$, then $W_{ij}=0$. The degree $d_i=  \sum_{j=1}^{\N} W_{ij}$ of a vertex $x_i$ is defined as the total weight of the edges linked to it.  The diagonal matrix $D$ with entries given by $D_{ii} = d_i$ is called the degree matrix.

A graph signal $f: V \rightarrow \R$ is a function that takes a real value on each vertex. A signal $f$ on a graph with $\N$ vertices can then be regarded as an $\N$-dimensional vector
$
 f=[f(x_1) \dots f(x_\N)]^T \in \R^\N.
$
The graph Laplacian matrix defined as
$
L=D-W
$
is of special importance in spectral graph theory \cite{Chung97, ShumanNFOV13}.
$L$ can be seen as an operator acting on the function $f$ through the matrix multiplication $Lf$, and it has been shown to be the graph equivalent of the Laplace operator in Euclidean domains, or the Laplace-Beltrami operator on manifold domains \cite{Chung97, HeinAL05, Singer06}. Recalling that the complex exponentials fundamental in classical signal processing have the special property that they are the eigenfunctions of the Laplace operator, one can extend the notion of frequency to graph domains. Relying on the analogy between the Laplace operator and the graph Laplacian $L$, one can define a Fourier basis on graphs, which consists of the eigenvectors of $L$. 

\begin{figure}[t]
\begin{center}
     \subfigure[]
       {\label{fig:u1}\includegraphics[height=2.7cm]{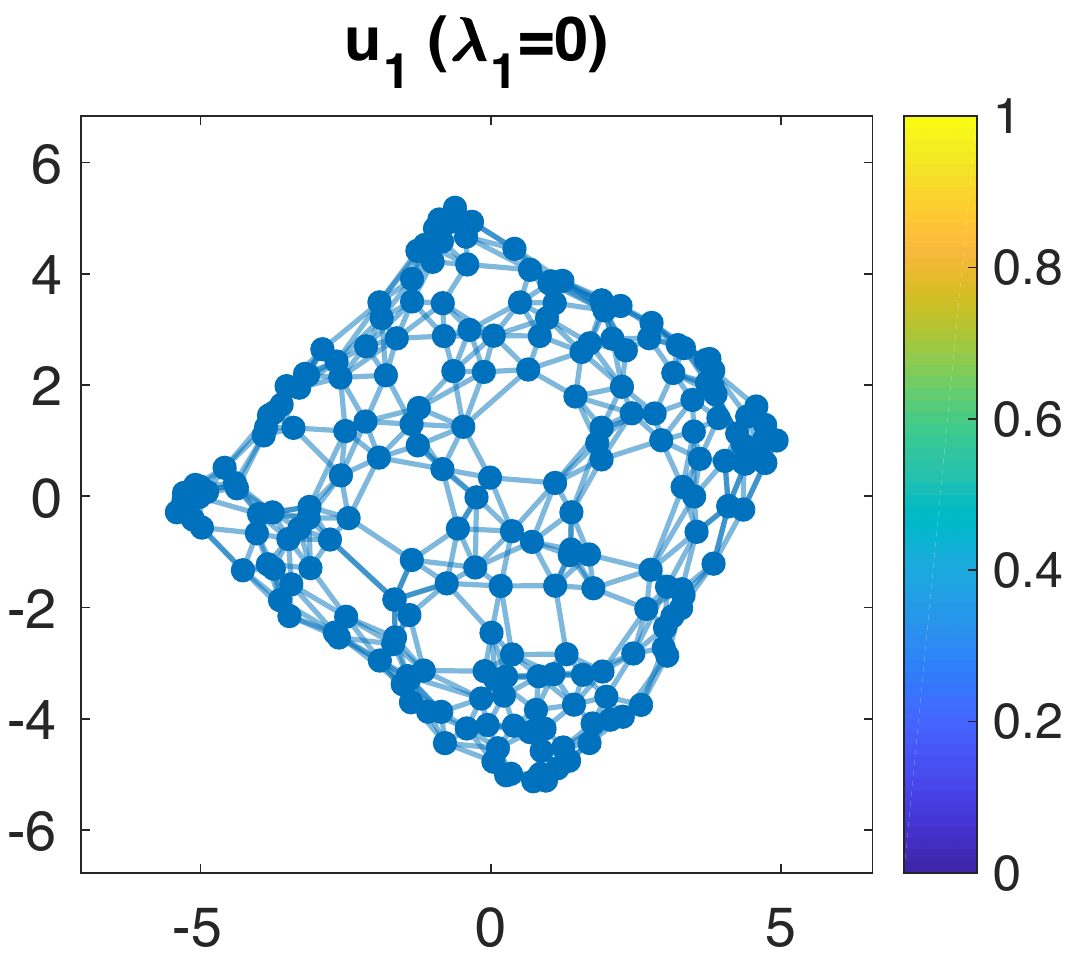}}
     \subfigure[]
       {\label{fig:u2}\includegraphics[height= 2.7cm]{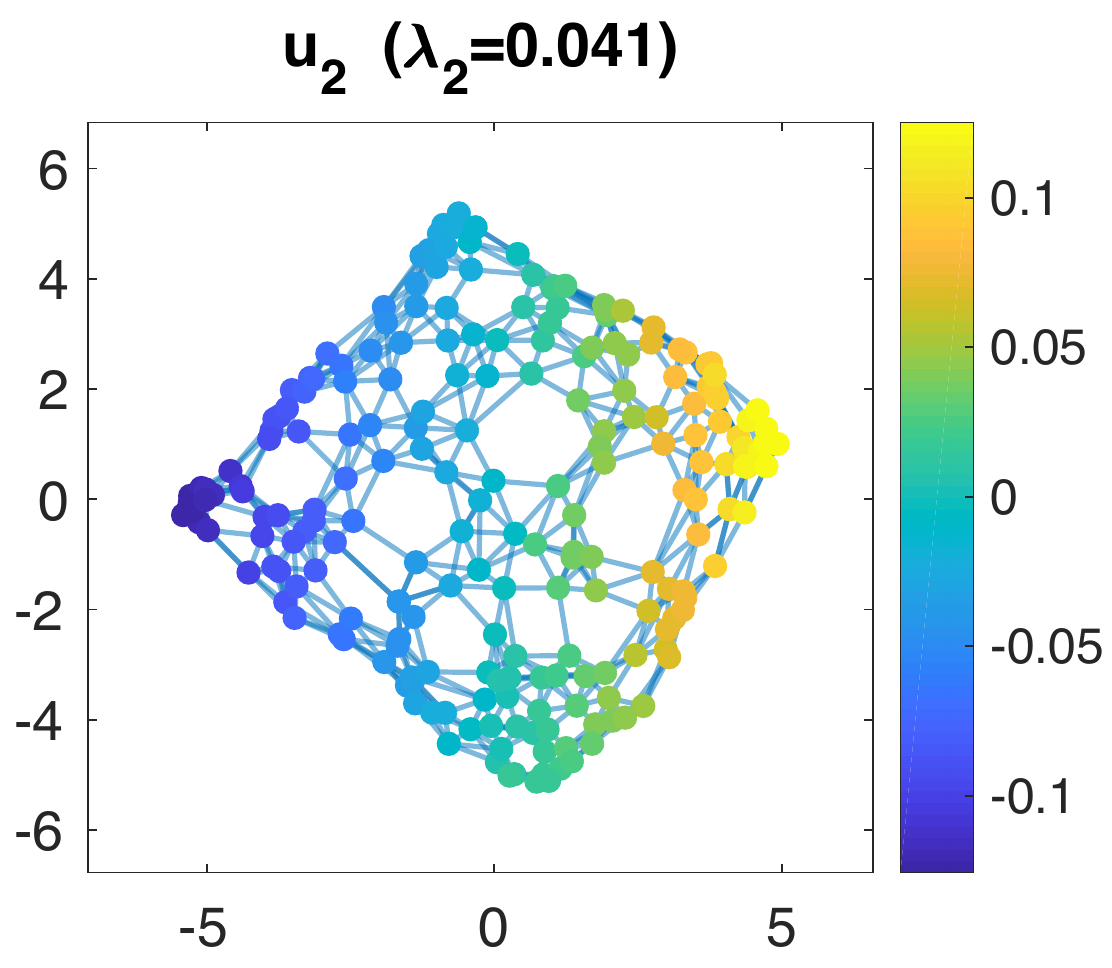}}           
     \subfigure[]
       {\label{fig:u3}\includegraphics[height= 2.7cm]{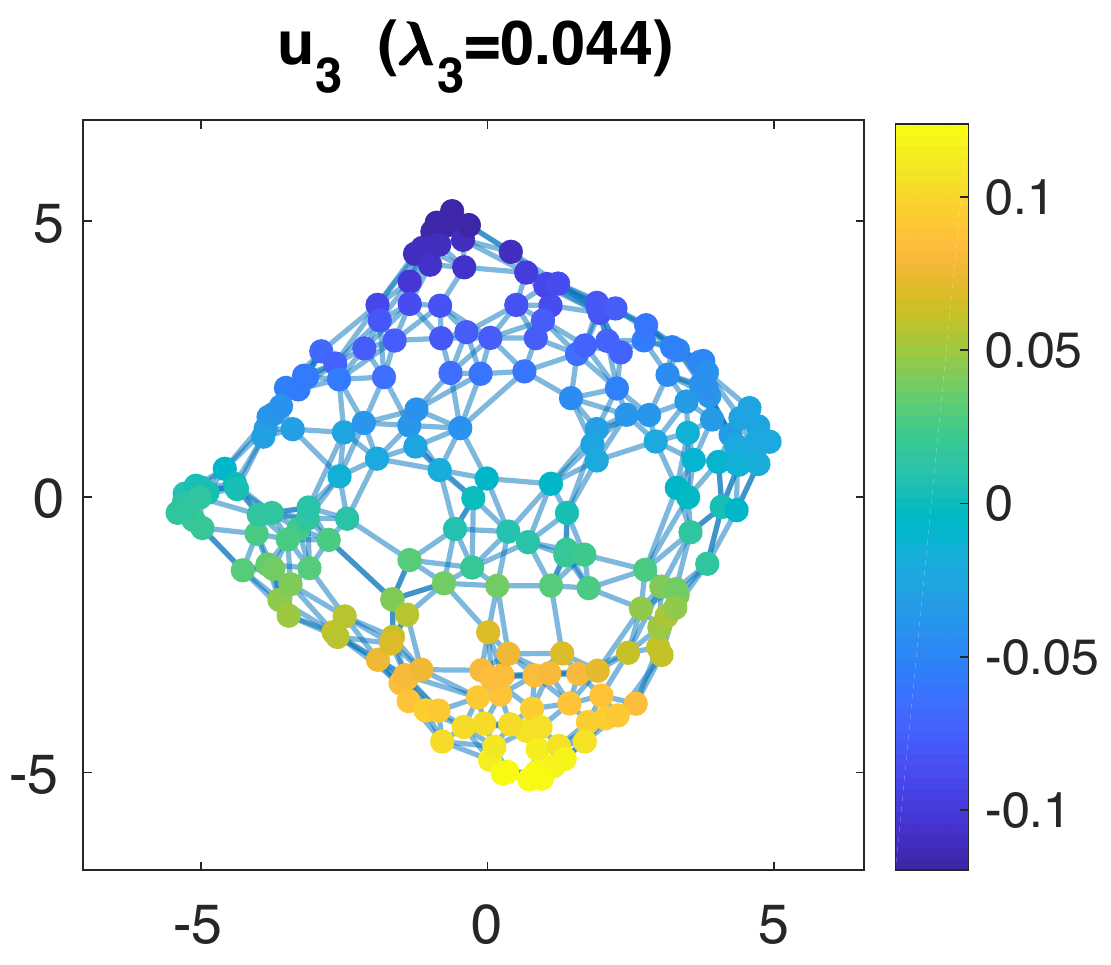}}
     \subfigure[]
       {\label{fig:u10}\includegraphics[height= 2.7cm]{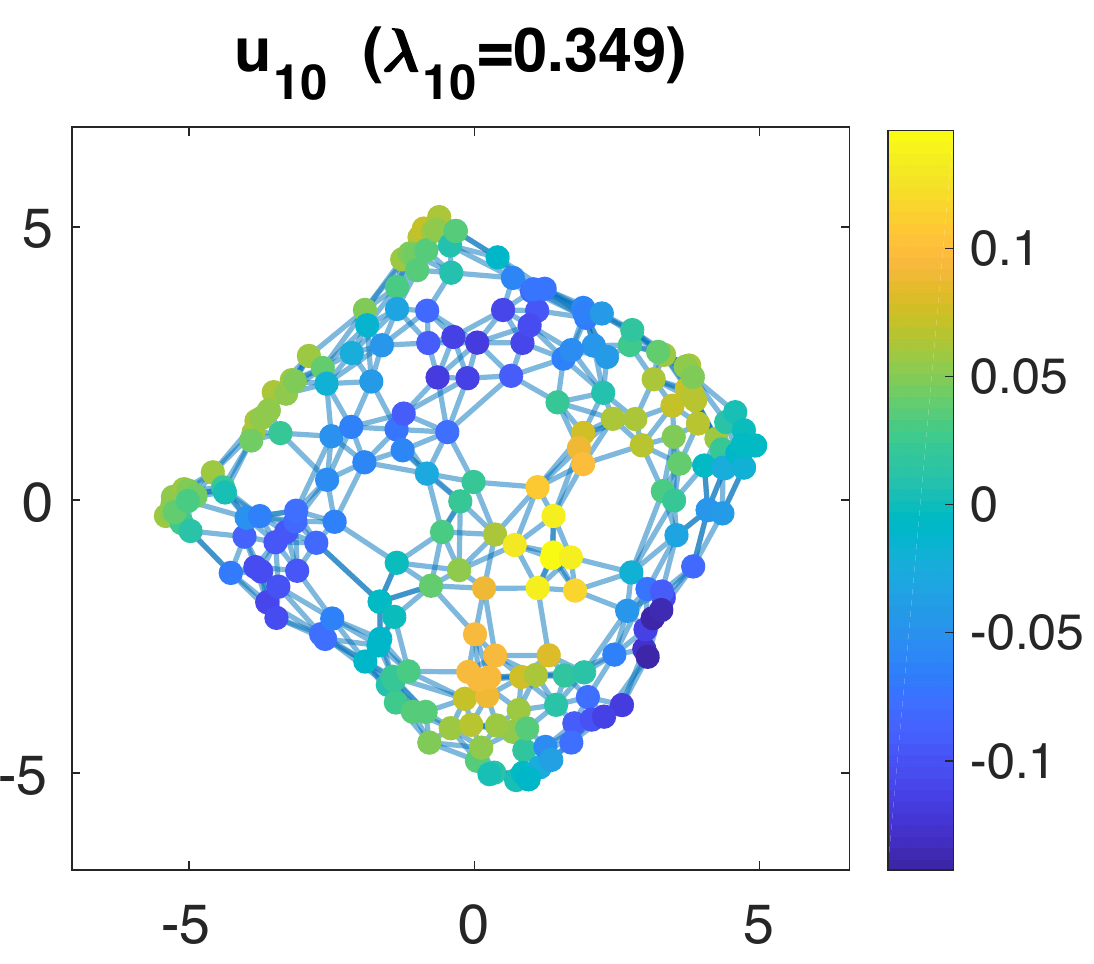}} 
      \subfigure[]
       {\label{fig:u50}\includegraphics[height= 2.7cm]{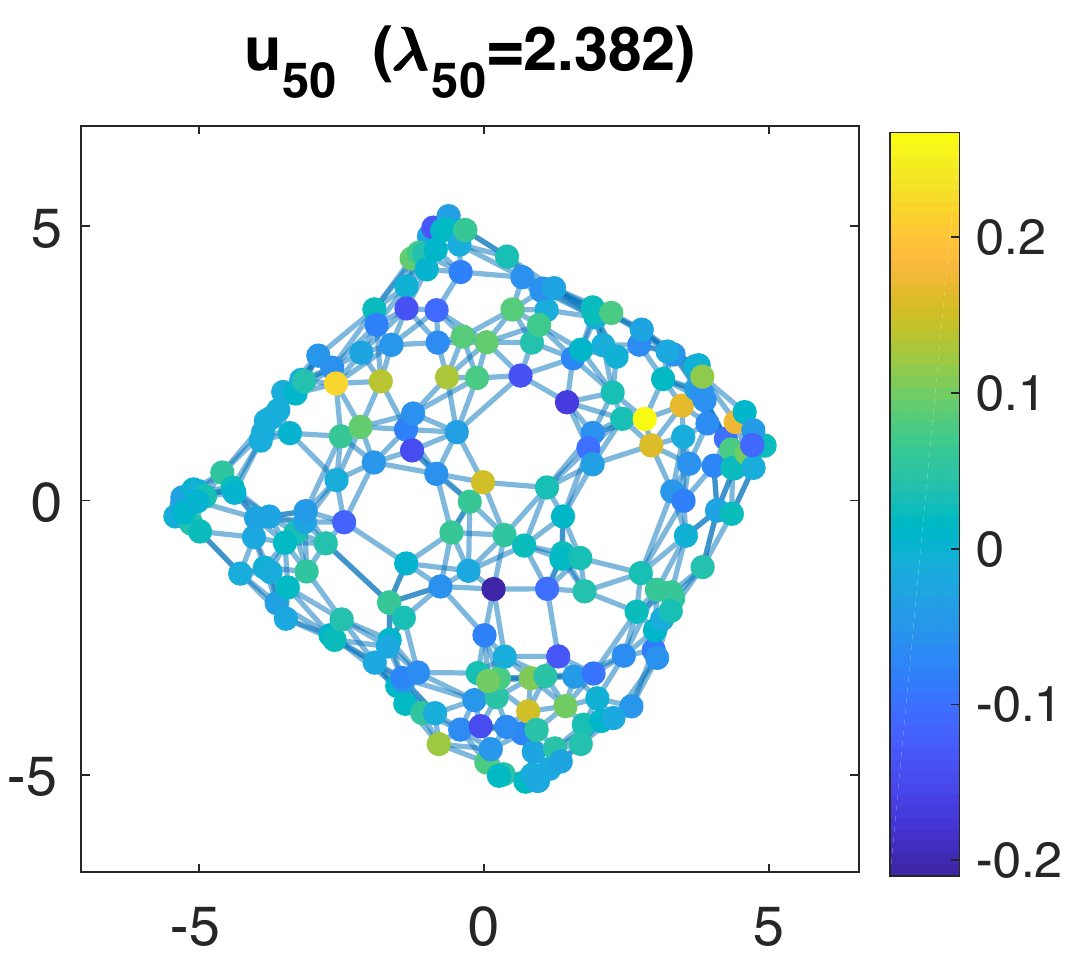}}       
 \end{center}
 \caption{Fourier basis vectors of an example graph. The eigenvectors $u_1$, $u_2$, $u_3$, $u_{10}$, $u_{50}$ of the graph Laplacian are plotted as graph signals in panels (a)-(e), where yellow and blue colors respectively indicate positive and negative values. The first Fourier basis vector $u_1$ has constant amplitude as its frequency is $\lambda_1=0$. The signals $u_2$ and $u_3$ have small frequencies around $0.04$ and they slowly oscillate along different directions on the graph. The signal $u_{10}$ has a larger frequency $\lambda_{10}=0.349$, hence its speed of oscillation is higher. Among the five signals, $u_{50}$ has the highest frequency $\lambda_{50}=2.382$ and it has the fastest variation on the graph.}
 \label{fig:illus_fourier_basis}
\end{figure}

Let $u_1, \dots, u_N$ denote the eigenvectors of the graph Laplacian, where
$
L u_k = \lambda_k u_k
$
for $k=1, \dots, N$. 
Here, each $u_k$ is a graph Fourier basis vector of frequency $\lambda_k$. The  eigenvector $u_1$ with the smallest eigenvalue $\lambda_1=0$ is always a constant function on the graph, and the speed of variation of $u_k$ on the graph  increases  for increasing $k$. The eigenvalues $\lambda_1, \dots, \lambda_N$ of the graph Laplacian correspond to frequencies such that $\lambda_k$ provides a measure of the speed of variation of the signal $u_k$ on the graph. The Fourier basis vectors of an example graph are illustrated in Figure \ref{fig:illus_fourier_basis}. In particular, the speed of variation of a signal $f$ over the graph is 
\[
f^T L f = \half \sum_{i,j=1}^\N W_{ij} (f(x_i) - f(x_j))^2,
\]
which takes larger values if the function $f$ varies more abruptly between neighboring graph vertices. Notice that the above term becomes the corresponding eigenvalue $\lambda_k$ of $L$  when $f$ is taken as $u_k$, since
$
u_k^T L u_k = \lambda_k.
$

This definition of the graph Fourier basis allows the extension of the Fourier transform to graph domains as follows. Let  $U=[u_1 u_2 \dots u_\N] \in \R^{\N \times \N}$ be the matrix consisting of the graph Fourier basis vectors. Then, for a graph signal $f \in \R^\N$, the Fourier transform of $f$ can simply be computed as
$
\alpha= U^T f,
$
where $\alpha = [\alpha_1 \dots \alpha_\N]^T$ is the vector consisting of the Fourier coefficients. Here $\alpha_k = u_k ^T  f$ is the $k$-th Fourier coefficient given by the inner product of $f$ and the Fourier basis vector $u_k$. Note that the graph Fourier basis $U$ is orthonormal as in classical signal processing; hence, the signal $f$ can be reconstructed from its Fourier coefficients as $f= U \alpha$.

\subsection{Notation and Setting}
\label{ssec:notation}

We now discuss the problem of domain adaptation on graphs and set the notation used in this paper. We consider a source graph $G^s=(V^s, E^s)$ with vertices $V^s=\{ x^s_i \}_{i=1}^{\Ns}$ and edges $E^s$; and a target graph $G^t=(V^t, E^t)$  with vertices $V^t=\{ x^t_i \}_{i=1}^{\Nt}$ and edges $E^t$. Let $W^s$ and $W^t$ denote the weight matrices, and let $L^s$ and $L^t$ be the Laplacians of the source and the target graphs. Let $f^s$ and $f^t$ be the label functions on the source and the target graphs, which represent class labels in a classification problem and continuously varying entities in a regression problem. We assume that some class labels are known as $ y^s_i = f^s(x^s_i) $ for $i \in \labS \subset \{ 1, \dots, \Ns \}$, and  $ y^t_i = f^t(x^t_i) $ for $i \in \labT \subset \{ 1, \dots, \Nt \}$, where $\labS $ and $\labT$ are index sets.  Many samples are labeled in the source domain and few samples are labeled in the target domain, i.e., $| \labT |  \ll  | \labS | $. Given the available labels  $\{y^s_i\}_{i \in  \labS }$ and $\{y^t_i\}_{i \in \labT }$, the purpose of graph domain adaptation is to compute accurate estimates $\hfs$, $\hft$ of $f^s$ and $f^t$.

All domain adaptation methods rely on a certain relationship between the source and the target domains. In this study, we consider a setting where a relationship can be established between the source and the target graphs through the frequency content of the label functions.
%
%
Let $f^s = U^s \alpha^s$ and $f^t = U^t \alpha^t $
denote the decompositions of the label functions over the source Fourier basis $U^s=[u_1^s \dots u_{\Ns}^s] \in \R^{\Ns \times \Ns}$ and the target Fourier basis $U^t=[u_1^t \dots u_{\Nt}^t] \in \R^{\Nt \times \Nt}$. We assume a setting where the source and the target label functions have similar spectra, hence, similar Fourier coefficients.

We have observed in our previous work \cite{PilanciV16} that, when computing the estimates $\hfs$ and $\hft$ of the label functions, it is useful to represent them in the reduced bases $\rU^s \in \mathbb{R}^{\Ns \times \sizeU }$ and $\rU^t \in \mathbb{R}^{\Nt \times \sizeU }$, which consist of the first $\sizeU$ Fourier basis vectors of smallest frequencies. This not only reduces the complexity of the problem but also has a regularization effect since components of very high frequency are excluded from the estimates. The estimates of $\fs$ and $\ft$ are then obtained in the form 
$
\hfs = \rU^s  \ralpha^s$ 
and
$
\hft = \rU^t  \ralpha^t,
$
where $\ralpha^s \in \R^{\sizeU}$ and $\ralpha^t \in \R^{\sizeU}$ are reduced Fourier coefficient vectors. In \cite{PilanciV16}, the label functions are estimated such that their Fourier coefficients $\ralpha^s$ and $\ralpha^t$ are close to each other and the estimates $\hfs$ and $\hft$ are consistent with the available labels.
 
\begin{figure}[t]
  \centering
  \centerline{\includegraphics[height=3cm]{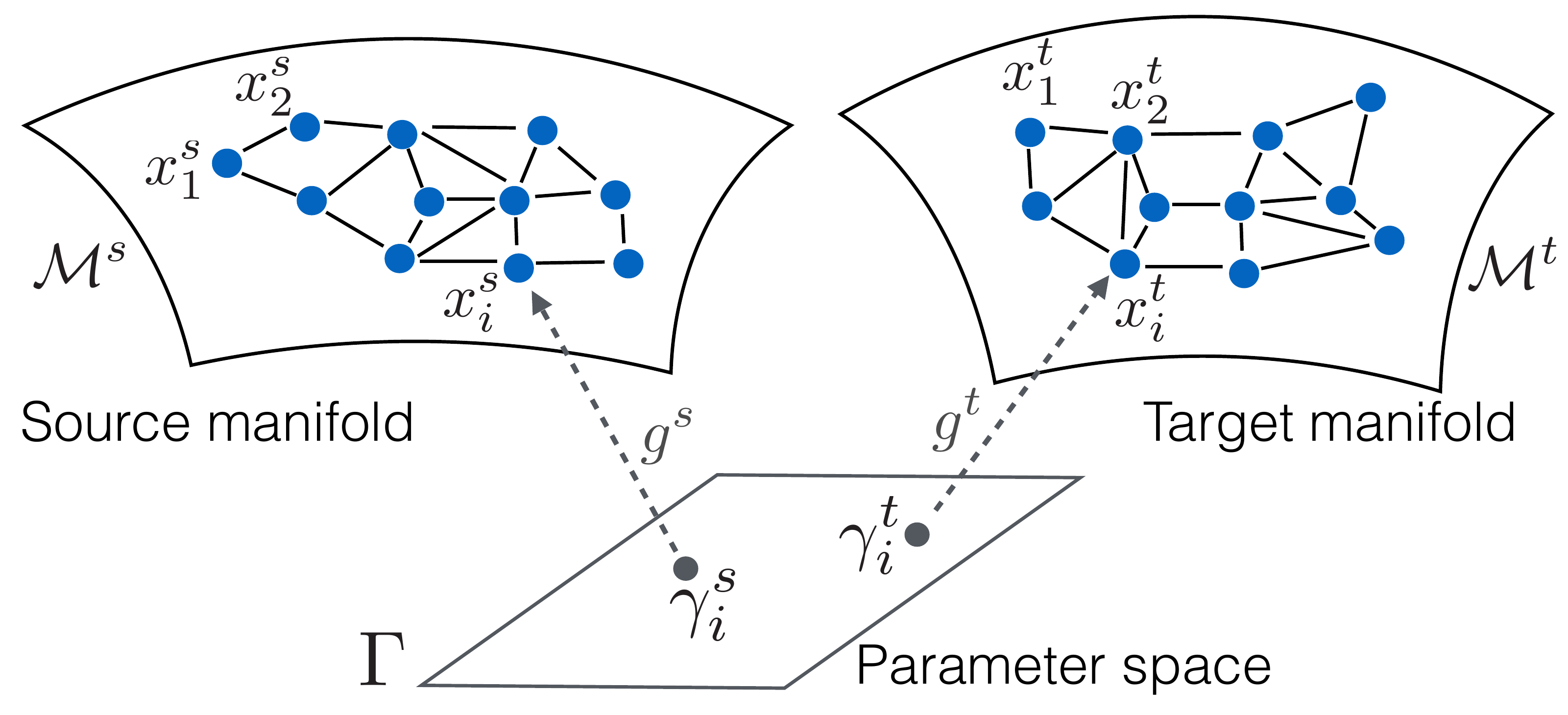}}
  \caption{Illustration of the domain adaptation setting considered in our study}
  \label{fig:illus_setting}
\end{figure}

In our theoretical analysis of graph domain adaptation, we consider a setting where graphs nodes are sampled from data manifolds. Let $\{ x^s_i \}_{i=1}^{\Ns} \subset \Ms$ and $\{ x^t_i \}_{i=1}^{\Nt} \subset \Mt$ denote source and target graph nodes sampled from a source data manifold $\Ms$ and a target data manifold $\Mt$. We assume that the source and the target data manifolds are generated through a pair of functions $\gs: \Gamma \rightarrow \Ms$ and $ \gt: \Gamma \rightarrow \Mt$ defined on a common parameter space $\Gamma$. Then, each manifold sample can be expressed as
$
x^s_i = \gs(\gamma^s_i)$ 
and
$
x^t_i = \gt(\gamma^t_i),
$
where $\gamma^s_i \in \Gamma$ and $\gamma^t_i \in \Gamma$ are parameter vectors as illustrated in Figure \ref{fig:illus_setting}. The parameter vectors are assumed to capture the source of variation generating the data manifolds. For instance, in a face recognition problem, $\gamma^s_i$ and $\gamma^t_i$ may represent rotation angles of the cameras viewing the subjects; and the discrepancy between $\Ms$ and $\Mt$ may result from the change in the illumination conditions. Note that in domain adaptation no relation is assumed to be known between $\gamma^s_i$ and $\gamma^t_i$. Moreover, the parameters $\{ \gamma^s_i  \}$, $\{ \gamma^t_i \}$ and the functions $\gs$, $\gt$ are often not known in practice. Although we consider this setting in our theoretical analysis, the practical algorithm we propose in Section \ref{sec:prop_method} will not require the knowledge of these parameters.

\subsection{Performance Bounds for Graph Domain Adaptation}

In this section, we analyze the error between the estimated target label function $\hft$ and the true target label function $\ft$. We would like to derive an upper bound for the estimation error
 \[ 
 E= \| \hft - \ft \|^2 = \sum_{i=1}^{\Nt}  ( \hft(x_i^t) - \ft(x_i^t) )^2 =  \sum_{i=1}^{\Nt} ( \hft_i - \ft_i )^2
\]
where $\hft_i = \hft(x_i^t)  $ and $\ft_i = \ft(x_i^t)  $ simply denote the values that the estimated and the true label functions take at the sample $x_i^t$.

We first define some parameters regarding the properties of the data manifolds. For the convenience of analysis, we assume that the source and the target graphs contain equally many samples\footnote{This assumption is made for simplifying the theoretical analysis. The algorithm proposed in Section \ref{sec:prop_method} does not require the source and the target graphs to have an equal number of nodes.}, i.e., $\Ns = \Nt = \N$.  We assume that the manifolds $\Ms \subset \Hs$ and $\Mt \subset \Ht$ are embedded in the Hilbert spaces $\Hs$, $\Ht$; the parameter space $\Gamma$ is a Banach space, and the manifolds $\Ms$ and $\Mt$ have (intrinsic) dimension $d$.

We assume that the functions $\gs: \Gamma \rightarrow \Ms$ and $\gt: \Gamma \rightarrow \Mt$ are Lipschitz-continuous, respectively with constants $\lips$ and $\lipt$; i.e., for any two parameter vectors $\gamma_1, \gamma_2 \in \Gamma$, we assume that
\[
\| \gs(\gamma_1) - \gs(\gamma_2) \| \leq  \lips  \|  \gamma_1 - \gamma_2 \|,
\qquad \qquad
\| \gt(\gamma_1) - \gt(\gamma_2) \| \leq  \lipt  \|  \gamma_1 - \gamma_2 \| 
\]
where $\| \cdot \| $ denotes the usual norm in the space of interest. Thus, the constants $\lips$ and $\lipt$ provide a measure of smoothness for the manifolds $\Ms$ and $\Mt$. We further assume that there exist two constants $\Al$ and $\Au$ such that for any $\gamma_1 \neq \gamma_2$ in $\Gamma$,
\begin{equation}
\label{defn:al_au}
\Al \leq  \frac{ \| \gt(\gamma_1) - \gt(\gamma_2)  \| }{ \| \gs(\gamma_1) - \gs(\gamma_2) \| } \leq  \Au.
\end{equation}
The constants $\Al$ and $\Au$ indicate the similarity between the geometric structures of the manifolds $\Ms$ and $\Mt$. As the variations of the functions $\gs$ and $\gt$ over $\Gamma$ become more similar, the constants $\Al$ and $\Au$ get closer to $1$. Let  
$
\A = \max(| 1- \Al |, |\Au-1|)
$
denote a bound on the deviations of $\Al$ and $\Au$ from $1$.

We consider a setting where the graph weight matrices are obtained with a kernel $\phi$ such that $W_{ij}^s =\phi(\| \xis - \xjs \|) $ and $W_{ij}^t =\phi(\| \xit - \xjt \|) $ for neighboring samples on the graph.
We assume that the kernel $\phi: \R^+ \cup \{  0\} \rightarrow \R^+$ is an $\Lphi$-Lipschitz nonincreasing function with
$
| \phi(u) - \phi(v)  | \leq \Lphi | u - v |
$
for any $u, v \in \R^+ \cup \{  0\} $. Let us denote the maximum value of the kernel function as $\phi_0 := \phi(0)$.

In our analysis, we consider domain adaptation algorithms that compute an estimate $\hft$ of $\ft$ whose values at the labeled nodes agree with the given labels, i.e.,
$ \hft_i = \ft_i $
for $i \in \labT$. Let $\wmin $, $K^{\max}$, and $K^{\min}$ be parameters representing the smallest edge weight, the maximum number of neighbors, and the minimum number of neighbors in the target graph. A more precise description of these parameters can be found in Appendix \ref{pf:thm_targetperf}.

We are now ready to state our main result in the following theorem.

\begin{theorem}
\label{thm:targetperf}

Consider a graph domain adaptation algorithm that estimates the source and the target label functions as $\hfs= \rU^s \ralpha^s$ and $\hft= \rU^t \ralpha^t$ such that the difference between their Fourier coefficients is bounded as $\| \ralpha^s - \ralpha^t \| \leq \dalpha$, the norms of the Fourier coefficients are bounded as $\| \ralpha^s \|, \| \ralpha^t \| \leq \calpha$, and $\hfs$ and $\hft$ are band-limited on the graphs so as not to contain any components with frequencies larger than $\lambdaR$. Assume that the estimate $\hfs$ is equal to the true source label function $\fs$ (e.g. as in a setting where all source samples are labeled).
Then, the target label estimation error can be upper bounded as
\begin{equation}
\label{eq:tar_est_error}
\| \hft  - \ft \|^2 \leq 
\frac{\kappa}{\wmin} (\sqrt{\B} + \sqrt{\hB})^2
\end{equation}
where
$\B$ is an upper bound on the rate of variation of the true label function
\begin{equation}
\label{eq:B_exp}
(\ft)^T L^t \ft \leq  \B,
\end{equation}
the parameter $\kappa $ is a function of the minimum and maximum number of neighbors of the form 
\begin{equation}
\label{eq:defn_kappa}
\kappa=O \left( 
\frac{1}{K^{\min}} \, 
\poly \left( \frac{K^{\max}}{K^{\min}}  \right )
\right),
\end{equation}
with $\poly(\cdot)$ denoting polynomial dependence, $\hB$ is an upper bound on the speed of variation of the target label estimate given by 
\begin{equation}
\label{eq:hatB_exp}
 (\hft)^T L^t \hft \leq  \hB := (\fs)^T L^s  \fs  +  \calpha^2 \rho_{\max} + 2 \calpha \lambdaR \dalpha,
\end{equation}
$\rho_{\max} $ is a geometry-dependent parameter varying at rate
\begin{equation}
\label{eq:rho_max_ord}
\rho_{\max} := O( \Lphi \, (\A  \, \lips + \lips + \lipt ) \egen  + \phi_0) ,
\end{equation}
and $\egen $ is proportional to the largest parameter-domain distance between neighboring graph nodes.

\end{theorem}

Theorem \ref{thm:targetperf} is stated more precisely in  Appendix \ref{pf:thm_targetperf}, where its proof can also be found. In the proof, first an upper bound  is derived on the difference between the rates of variation of the source and the target label functions. Next, the deviation between the eigenvalues of the source and the target graph Laplacians is studied. Finally, these two results are combined to obtain an upper bound on the estimation error of the target label function. 
%
%

Theorem \ref{thm:targetperf} can be interpreted as follows. First, observe that the estimation error increases linearly with the bound $\dalpha $ on the deviation between the source and the target Fourier coefficients. This suggests that in graph domain adaptation it is favorable to estimate the source and the target label functions in a way that they have similar spectra. The theorem also has several implications regarding the smoothness of label functions and their estimates. It is well known that graph-based learning methods perform better if label functions vary smoothly on the graph. This is formalized in Theorem \ref{thm:targetperf} via the assumption that the estimates $\hfs$ and $\hft$ are band-limited such that the highest frequency present in their spectrum (computed with the graph Fourier transform) does not exceed some threshold $\lambdaR $, which limits their speeds of variation on the graphs. Notice that the rates of variation $(\fs)^T L^s  \fs $ and  $ (\ft)^T L^t  \ft $ of the true source and target label functions also affect the estimation error through the terms $\B$ and $\hB$.

Next, we observe from \eqref{eq:tar_est_error} that the estimation error depends on the geometric properties of data manifolds as follows. The error increases linearly with $\rho_{\max}$ (through the term $\hB$), while in \eqref{eq:rho_max_ord}, $\rho_{\max}$ is seen to depend linearly on the parameters $\A$,  $\lips$, and $\lipt$. Recalling that $\lips$ and $\lipt$ are the Lipschitz constants of the functions $\gs$ and $\gt$ defining the data manifolds, the theorem suggests that the estimation error is smaller when the data manifolds are smoother. The fact that the error increases linearly with the parameter $\A$ is intuitive as $\A$ captures the dissimilarity between the geometric structures of the source and the target manifolds. We also notice that $\rho_{\max}$ is proportional to the parameter $\egen$. We give a precise definition of the parameter $\egen$ in Appendix \ref{pf:thm_targetperf}, which can be roughly described as an upper bound on the parameter-domain ($\Gamma$) distance between neighboring graph samples. As the number of samples $\N$ increases, $\egen$ decreases at rate $ \egen = O ( \N^{-1/d} ) $, where $d$ is the intrinsic dimension of the manifolds. Since $\rho_{\max}$ is linearly proportional to $\egen$, we conclude that the estimation error of the target label function decreases with $\N$ at the same rate $ O (  \N^{-1/d} ) $. This can be intuitively interpreted in the way that the discrepancy between the topologies of the source and the target graphs resulting from finite sampling effects decreases as the sampling of the data manifolds becomes denser.

The result in Theorem \ref{thm:targetperf} also leads to the following important conclusions about the effect of the graph properties on the performance of learning. First, the estimation error is observed to increase linearly with the parameter-domain distance $\egen$ between neighboring points on the graphs. This suggests that when constructing the graphs, two samples that are too distant from each other in the parameter space should rather not be connected with an edge. Then, we notice that the parameter $\kappa$ decreases when the ratio $K^{\max} / K^{\min}$ between the maximum and the minimum number of neighbors is smaller. Hence, more ``balanced'' graph topologies influence the performance positively; more precisely, the number of neighbors of different graph nodes should not be disproportionate. At the same time,  the term $K^{\min}$ in the denominator in \eqref{eq:defn_kappa} implies that nodes with too few neighbors should rather be avoided. A similar observation can be made about the term $\wmin$ in the expression of the error bound in \eqref{eq:tar_est_error}. The minimal edge weight term $\wmin$ in the denominator suggests that graph edges with too small weights have the tendency to increase the error. From all these observations, we conclude that when constructing graphs, balanced graph topologies should be preferred and significant variation of the number of neighbors across different nodes, too isolated nodes, and too weak edges should be avoided.

\section{Learning Graph Topologies for Domain Adaptation}
\label{sec:prop_method}

In this section, we propose an algorithm for jointly learning graphs and label functions for domain adaptation, based on the theoretical findings presented in Section \ref{sec:theo_analy}. We first formulate the graph domain adaptation problem and then propose a method for solving it.

\subsection{Problem Formulation}

Given the source and the target samples $\{ \xis \}_{i=1}^{\Ns} \subset \R^n$ and $\{\xit \}_{i=1}^{\Nt} \subset \R^n$, we consider the problem of constructing a source graph and a target graph with respective vertex sets $\{ \xis \}$ and $\{ \xit \}$, while obtaining estimates $\hfs= \rU^s \ralpha^s$ and $\hft= \rU^t \ralpha^t$ of the label functions at the same time. The problem of learning the graph topologies is equivalent to the problem of learning the weight matrices $W^s$ and $W^t$. 

The bound \eqref{eq:tar_est_error} on the target error suggests that when learning a pair of graphs, the parameters $\kappa$, $\B$  and $\hB$ should be kept small, whereas small values for $\wmin$ should be avoided. The expression in \eqref{eq:hatB_exp} shows that the parameters $\lambdaR$ and $\dalpha$ should be kept small. Meanwhile, in the expression of $\rho_{\max}$ in \eqref{eq:rho_max_ord}, we observe that the terms $\A$, $\lips$, and $\lipt$  are determined by the geometry of the data manifolds and these are independent of the graphs. On the other hand, the parameter $\egen$ depends on the graph topology and can be controlled more easily. Thus, in view of the interpretation of Theorem \ref{thm:targetperf}, we propose to jointly learn the label functions $\hfs$ and $\hft$ and the weight matrices $W^s$ and $W^t$ based on the following optimization problem.

\begin{equation}
\label{eq:main_opt_prob}
\begin{split}
&\text{minimize}_{\ralpha^s, \ralpha^t, W^s, W^t}  \   \|  S^s \rU^s \ralpha^s -  y^s \|^2 + \|  S^t \rU^t \ralpha^t -  y^t \|^2 + \mu \| \ralpha^s - \ralpha^t  \|^2 \\
& \qquad \qquad \qquad \qquad 
 +  (\hfs)^T L^s  \hfs  +  (\hft)^T L^t  \hft 
+ \mu_s \sum_{i,j=1}^{\Ns} W^s_{ij} \|  \xis - \xjs \|^2 + \mu_t \sum_{i,j=1}^{\Nt} W^t_{ij} \|  \xit - \xjt \|^2 \\
& \text{subject to }  W^s_{ij} \geq W^{\min}, \ \forall i,j \in \{1, \dots, \Ns \} \text{ with } W^s_{ij} \neq 0; 
 \quad
 \ W^t_{ij} \geq  W^{\min},  \ \forall i,j  \in \{1, \dots, \Nt \} \text{ with } W^t_{ij} \neq 0;  \\
 & \qquad  \qquad
  \dmin \leq d^s_i  \leq \dmax, \ \forall i \in \{1, \dots, \Ns \} ; \quad
  \dmin \leq d^t_i  \leq \dmax, \ \forall i \in \{1, \dots, \Nt \} .
\end{split}
\end{equation}
Here  $\mu$, $\mu_s$, and $\mu_t$ are positive weight parameters, and $y^s$ and $y^t$ are vectors consisting of all the available source and target labels. The first two terms $\|  S^s \rU^s \ralpha^s -  y^s \|^2$ and $\|  S^t \rU^t \ralpha^t -  y^t \|^2$ in \eqref{eq:main_opt_prob} enforce the estimated label functions $\hfs$ and $\hft$ to be consistent with the available labels, where $S^s$ and $S^t$ are binary selection matrices consisting of $0$'s and $1$'s that select the indices of labeled data. The third term $\| \ralpha^s - \ralpha^t  \|^2$ aims to reduce the parameter $\dalpha$ in \eqref{eq:hatB_exp}. Note that the representation of $\hfs$ and $\hft$ in terms of the first $\sizeU$ Fourier basis vectors in $\rU^s$ and $\rU^t $ is useful for keeping the parameter $\lambdaR$ small.  

Next, the minimization of the terms $(\hfs)^T L^s  \hfs$ and $(\hft)^T L^t  \hft $ encourages the label functions $\hfs$ and $\hft$ to vary slowly on the graphs, which aims to reduce the parameters $\hB$ and $\B$  in \eqref{eq:hatB_exp} and \eqref{eq:B_exp}. We also recall from Theorem \ref{thm:targetperf} that in order to make the parameter $ \egen$ small, graph edges between distant points should be avoided. The terms $ \sum_{i,j} W^s_{ij} \|  \xis - \xjs \|^2$ and $ \sum_{i,j} W^t_{ij} \|  \xit - \xjt \|^2 $ aim to achieve this by penalizing large edge weights between distant samples. The inequality constraints $ W^s_{ij} \geq W^{\min} $ and $ W^t_{ij} \geq  W^{\min}$ on nonzero edge weights ensure that the minimum edge weight $\wmin$ is above some predetermined threshold $W^{\min}$. 

Finally, we recall from Theorem \ref{thm:targetperf} that in order to minimize $\kappa$, the ratio $K^{\max} / K^{\min}$ must be kept small while avoiding too small $K^{\min}$ values. However, incorporating the number of neighbors directly in the objective function would lead to an intractable optimization problem. Noticing that the node degrees are expected to be proportional to the number of neighbors, we prefer to relax this to the constraints  $\dmin \leq d^s_i  \leq \dmax$ and $ \dmin \leq d^t_i  \leq \dmax$ on the node degrees, where $d^s_i$ and $d^t_i$ respectively denote the degrees of $\xis$ and $\xit$; and $\dmin$ and $\dmax$ are some predefined degree threshold parameters with $0<\dmin \leq \dmax$.

\subsection{Proposed Method: Domain Adaptive Graph Learning}

Analyzing the optimization problem in \eqref{eq:main_opt_prob}, we observe that the matrices $\rU^s$ and $\rU^t$ are nonconvex and highly nonlinear functions of the optimization variables $W^s$ and $W^t$ as they consist of the eigenvectors of the graph Laplacians $L^s$ and $L^t$. Moreover, due to the multiplicative terms such as $\rU^s \ralpha^s$, the problem is even not jointly convex in $\ralpha^s$, $\ralpha^t$, $\rU^s$, and $\rU^t$. Hence, it is quite difficult to solve the problem \eqref{eq:main_opt_prob}. In our method, we employ a heuristic iterative solution approach that relaxes the original problem \eqref{eq:main_opt_prob} into more tractable subproblems and alternatively updates the coefficients and the weight matrices in each iteration as follows.

We first initialize the weight matrices $W^s$, $W^t$ with a typical strategy; e.g., by connecting each node to its $K$ nearest neighbors and assigning edge weights with a Gaussian kernel. We use the normalized versions of the graph Laplacians given by $L^s= (D^s)^{-1/2} (D^s - W^s)(D^s)^{-1/2} $ and $L^t= (D^t)^{-1/2} (D^t - W^t)(D^t)^{-1/2}$. 

In the first step of each iteration, we optimize $\ralpha^s$, $\ralpha^t$ by fixing the weight matrices $W^s$, $W^t$. This gives the following optimization problem:
\begin{equation}
\label{eq:obj_SDA}
\text{minimize}_{\ralpha^s, \ralpha^t}    
\|  S^s \rU^s \ralpha^s -  y^s \|^2 + \|  S^t \rU^t \ralpha^t -  y^t \|^2 + \mu \| \ralpha^s - \ralpha^t  \|^2 .
\end{equation}
The simplified objective\footnote{Note that the dependence of $\hfs$ and $\hft$ on $\ralpha^s$ and $\ralpha^t$ is neglected in \eqref{eq:obj_SDA}. The reason is that since $\rU^s$ and $\rU^t$ consist of the eigenvectors of $L^s$ and $L^t$, the terms $(\hfs)^T L^s  \hfs$ and $(\hft)^T L^t  \hft $ would contribute to the objective only as regularization terms on the weighted norms  of $\ralpha^s$ and $\ralpha^t$. We prefer to exclude such a regularization in order to prioritize fitting the coefficients $\ralpha^s$, $\ralpha^t $ to each other and to the available labels.} in \eqref{eq:obj_SDA} is in fact the same as the objective of the SDA algorithm proposed in \cite{PilanciV16}. As the problem is quadratic and convex in $\ralpha^s$ and $\ralpha^t$, its solution can be analytically found by setting the gradient equal to $0$, which gives \cite{PilanciV16}
\begin{equation*}
\ralpha^s = ( \mu^{-1}  A^t A^s + A^t + A^s  )^{-1}  
		(    \mu^{-1}  A^t  B^s y^s +   B^s y^s + B^t y^t   ),
\qquad \qquad
\ralpha^t =  (	\mu^{-1} A^s \ralpha^s + \ralpha^s 	 - \mu^{-1}  B^s y^s  )
\end{equation*}
where
$A^s = (\rU^s)^T (S^s)^T  S^s \rU^s $,  
$\quad B^s = (\rU^s)^T (S^s)^T$, 
$\quad A^t = (\rU^t)^T (S^t)^T S^t \rU^t $,
$ \quad B^t = (\rU^t)^T (S^t)^T$.

\begin{algorithm}[t]
\footnotesize
\caption{Spectral Domain Adaptation via Domain Adaptive Graph Learning (SDA-DAGL)}

\begin{algorithmic}[1]

\STATE
\textbf{Input:} \\
$\{ \xis \}$, $\{ \xit \}$: Source and target samples\\
$y^s$, $y^t$: Available source and target labels\\

\STATE
\label{alg:state_init_sdadagl}
\textbf{Initialization:}

Initialize the weight matrices $W^s$, $W^t$ with a sufficiently large number of edges, e.g., as K-NN graphs.

\REPEAT

\STATE
Compute the graph Laplacians $L^s$, $L^t$ and Fourier bases $\rU^s$, $\rU^t$ with weight matrices $W^s$, $W^t$.

\STATE
\label{alg:state_coef_upd}
 Update coefficients $\ralpha^s$, $\ralpha^t$ by solving \eqref{eq:obj_SDA}.
 
\STATE
\label{alg:state_weight_upd}
Update the weight matrices $W^s$, $W^t$ by solving \eqref{eq:opt_LP_weights} via linear programming.

\STATE
Prune the graph edges by setting the edge weights with $W^s_{ij} < W^{\min}$ and $W^t_{ij} < W^{\min}$ to 0.

\UNTIL the maximum number of iterations is attained

\STATE
\textbf{Output}:\\

$f^t =  \rU^t  \ralpha^t $: Estimated target label function\\
$f^s = \rU^s  \ralpha^s$: Estimated source label function

\end{algorithmic}
\label{alg:SDA_DAGL_Algorithm}
\end{algorithm}

Then, in the second step of an iteration, we fix the coefficients $\ralpha^s$ and $\ralpha^t$, and optimize the weight matrices $W^s$ and $W^t$. As the dependence of the Fourier basis matrices $\rU^s$ and $\rU^t$ on $W^s$ and $W^t$ is quite intricate, we fix $\rU^s$ and $\rU^t$ to their values from the preceding iteration and neglect this dependence when reformulating our objective for learning the weight matrices. Defining the vectors $\hhs=(D^s)^{-1/2} \hfs$ and $\hht=(D^t)^{-1/2} \hft$, the fourth and fifth terms in \eqref{eq:main_opt_prob} can be rewritten as 
\[
 (\hfs)^T L^s  \hfs  +  (\hft)^T L^t  \hft  
 = (\hhs)^T (D^s -  W^s) \hhs  +  (\hht)^T (D^t -  W^t) \hht.
\]
If the node degrees were fixed, the minimization of the above term would correspond to the maximization of $(\hhs)^T W^s \hhs  +  (\hht)^T  W^t \hht$.  However, we have observed that letting the node degrees vary in an appropriate interval gives better results than fixing them. We thus propose the problem  
\begin{equation}
\label{eq:opt_LP_weights}
\begin{split}
\text{minimize}_{\ W^s, W^t}  \
&  \   \mu_s \sum_{i,j=1}^{\Ns} W^s_{ij} \|  \xis - \xjs \|^2  -  (\hhs)^T W^s \hhs
+ \mu_t \sum_{i,j=1}^{\Nt} W^t_{ij} \|  \xit - \xjt \|^2   -  (\hht)^T  W^t \hht \\
 \text{subject to } 
&  \dmin \leq  \sum_{j=1}^{\Ns} W^s_{ij}  \leq \dmax, \ \forall i = 1, \dots, \Ns  ;  \quad
0 \leq W^s_{ij}  \leq 1, \, \forall i, j = 1, \dots, \Ns;   \\
&  \dmin \leq  \sum_{j=1}^{\Nt} W^t_{ij}    \leq \dmax, \ \forall i = 1, \dots, \Nt ; \quad
  0 \leq W^t_{ij}  \leq 1, \, \forall i, j = 1, \dots, \Nt.
\end{split}
\end{equation}
for optimizing $W^s$ and $W^t$. The objective function and the constraints of the problem \eqref{eq:opt_LP_weights} are linear in the entries of $W^s$ and $W^t$. Hence, \eqref{eq:opt_LP_weights} can be posed as a linear programming (LP) problem and can be solved with an LP solver. 
 In the problem in \eqref{eq:opt_LP_weights}, the edge weights are constrained to lie between 0 and 1.  Since the solution of an LP problem occurs at a corner point of the feasible region, many entries of the weight matrices solving the LP problem in \eqref{eq:opt_LP_weights} are 0. This gradually improves the sparsity of the weight matrices. Then, instead of directly incorporating the sparsity of the weight matrices in the optimization problem and imposing a lower bound on the positive edge weights as in the original problem  \eqref{eq:main_opt_prob},  we prefer to initialize the graphs with a sufficiently high number of edges, solve the LP problem   \eqref{eq:opt_LP_weights} by optimizing only the nonzero edge weights, and then at the end of each iteration apply a graph pruning step that sets the edge weights smaller than $W^{\min}$ to 0. 
After each iteration, the graph Laplacians $L^s$, $L^t$ and the Fourier bases $\rU^s$, $\rU^t$ are updated. The same procedure then continues with the optimization of the Fourier coefficients as in \eqref{eq:obj_SDA}. We call this algorithm Spectral Domain Adaptation via Domain Adaptive Graph Learning (SDA-DAGL) and summarize it in Algorithm \ref{alg:SDA_DAGL_Algorithm}. Due to the various relaxations made in different stages, it is not possible to guarantee the convergence of the solution in general. In practice, we have found it useful to terminate the algorithm after a suitably chosen number of iterations.

\section{Experimental Results}
\label{sec:exp_res}

\begin{figure}[t]
\begin{center}
     \subfigure[Synthetic data set]
       {\label{fig:toydata}\includegraphics[height=3cm]{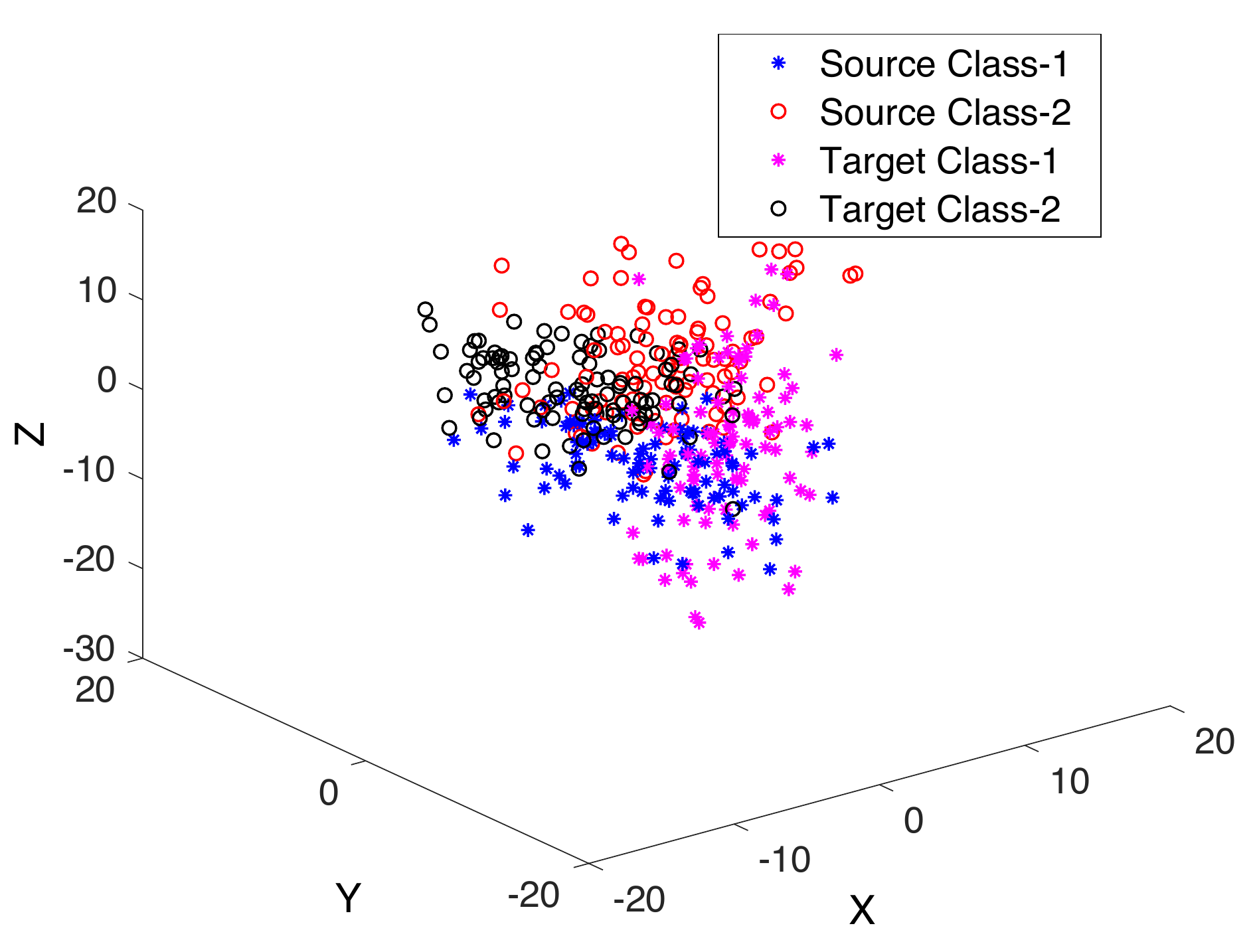}}
      \subfigure[COIL-20 data set]
       {\label{fig:Coil_data_matches}\includegraphics[width=8cm]{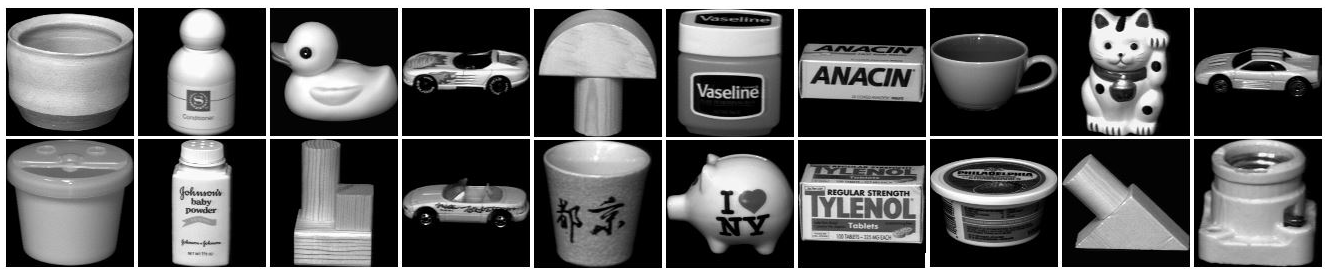}}
 \end{center}
 \vspace{-0.4cm}
 \caption{(a) Synthetic data set with two classes drawn from normal distributions. (b) Sample images from the COIL-20 data set. Each source domain object in the upper row is assigned the same class label as the matching target domain object right below it.}
 \label{fig:datasets}
 \vspace{-0.4cm}
\end{figure}

We now evaluate the proposed method with experiments on a synthetic data set and a real data set. The synthetic data set shown in Figure \ref{fig:toydata} consists of 400 normally distributed  samples in  $\R^3$ from two classes. The two classes in each domain have different means, and the source and the target domains differ by a rotation of $90^\circ$. The variance of the normal distributions is chosen to be relatively large to ensure a sufficient level of difficulty.  The COIL-20 object database \cite{NeneNM96} shown in Figure \ref{fig:Coil_data_matches} consists of a total of 1440 images of 20 objects. Each object has 72 images taken from different camera angles rotating around it. We downsample the images to a resolution of $32 \times 32$ pixels. The 20 objects in the data set are divided into two groups and each object in the first group is matched to the object in the second group that is the most similar to it. Each group of 10 objects is taken as a different domain and the matched object pairs are considered to have the same class label in the experiments.

\begin{figure}[t]
\begin{center}
     \subfigure[Synthetic data set]
       {\label{fig:error_vs_K_toy}\includegraphics[height=3.5cm]{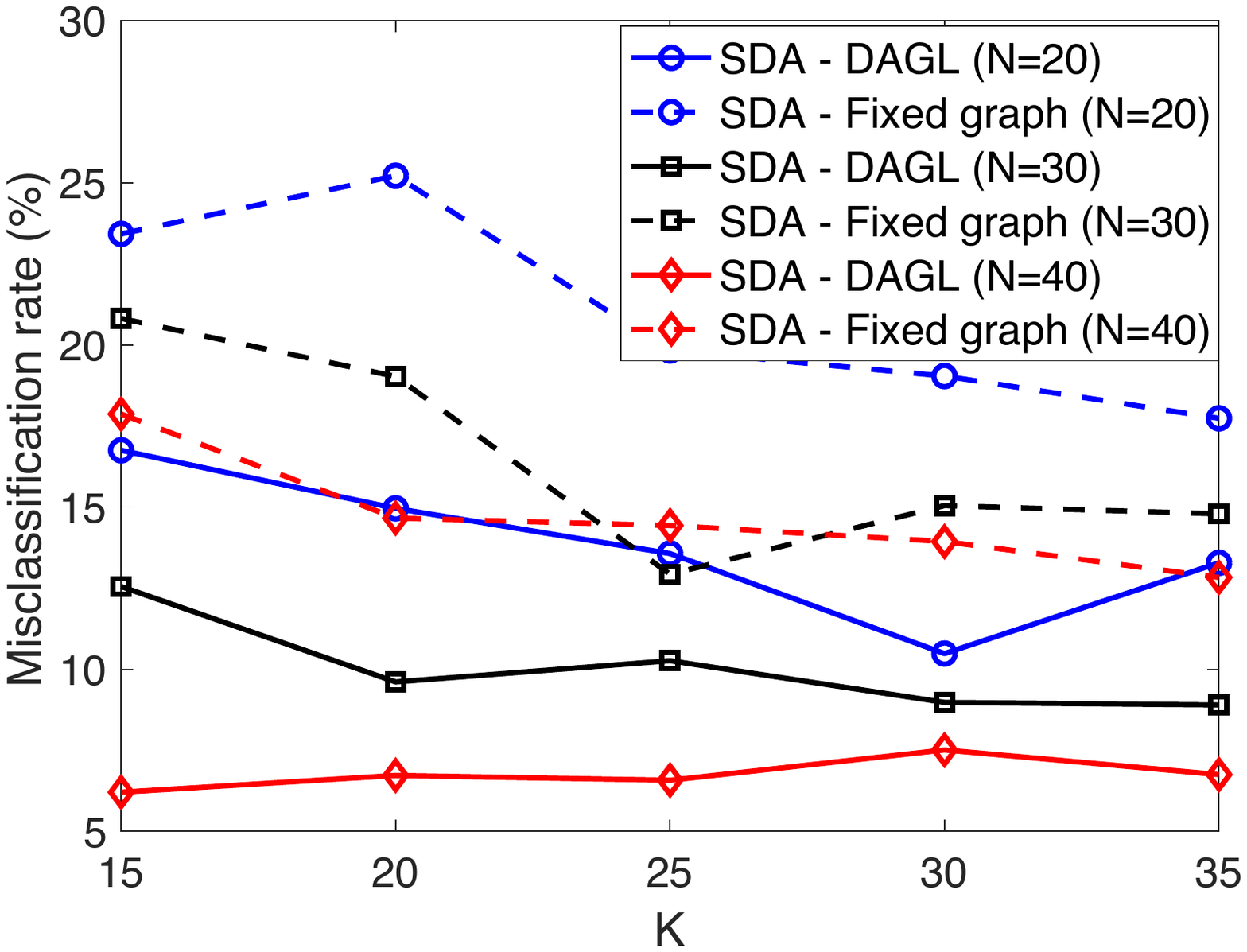}}
      \subfigure[COIL-20 data set]
       {\label{fig:error_vs_K_coil}\includegraphics[height=3.5cm]{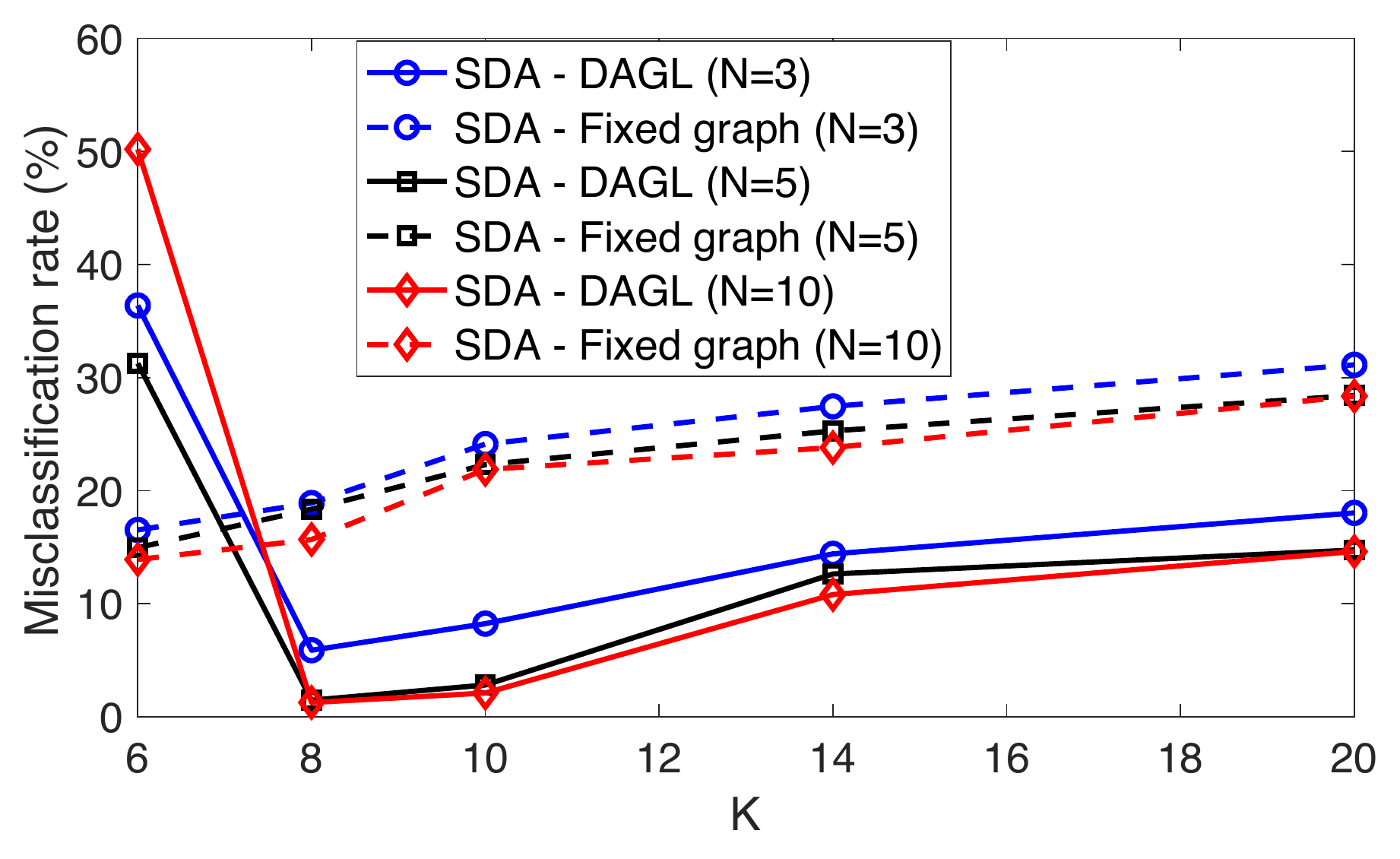}}
        \subfigure[Target weight matrix]
       {\label{fig:Wt_iter}\includegraphics[height=4.0cm]{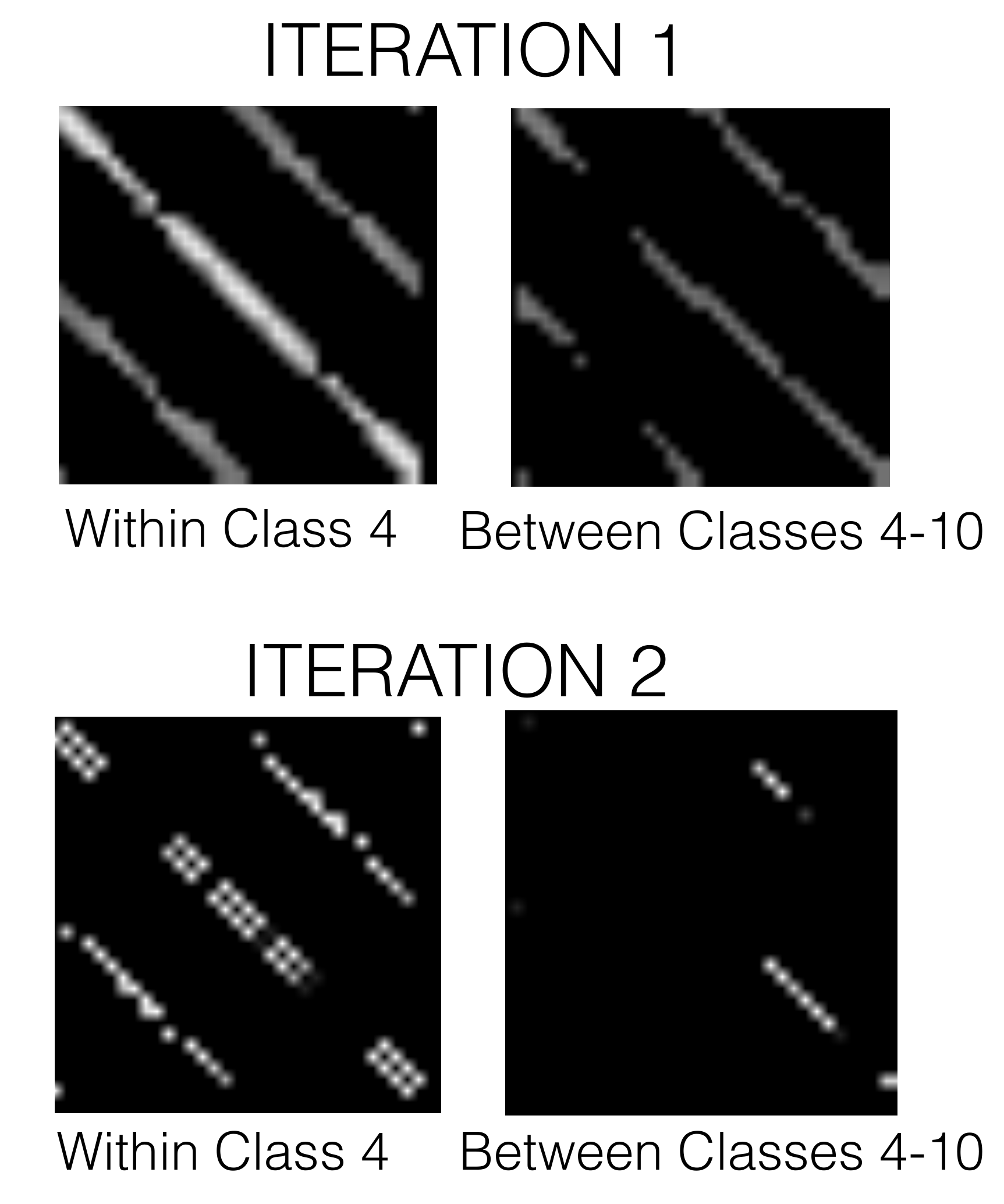}}
 \end{center}
 \vspace{-0.5cm}
 \caption{(a), (b) The variation of the misclassification rates of the target samples with the number of nearest neighbors  $K$. The curves with dashed lines are obtained with a fixed graph topology (SDA), whereas the corresponding curves with solid lines are obtained with the proposed graph learning method (SDA-DAGL). The results show that the graphs learnt with the proposed method yield higher performance than fixed graph topologies constructed with K-NN. (c) The evolution of $W^t$ during two consecutive iterations. The black color represents 0 weight (no edge) and brighter tones indicate larger edge weights. The updates in the graph topology tend to preserve within-class edges and suppress between-class edges as desired.}
 \label{fig:err_vs_K}
 \vspace{-0.4cm}
\end{figure}

We first compare our SDA-DAGL algorithm with the SDA algorithm in order to study the efficiency of the proposed graph learning approach. In both data sets, we first  independently construct the source and the target graphs by connecting each sample to its $K$ nearest neighbors and forming the weight matrices $W^s$ and $W^t$ with a Gaussian kernel. The SDA algorithm uses the fixed graph topology represented by these weight matrices. In the proposed SDA-DAGL method, these weight matrices are used to initialize the algorithm as in Step \ref{alg:state_init_sdadagl} of Algorithm \ref{alg:SDA_DAGL_Algorithm}, and then they are refined gradually with the proposed joint graph learning and label estimation approach. All class labels are known in the source domain, while a small number of labels are known in the target domain. The class labels of the unlabeled target samples are estimated with the two algorithms in comparison. Figure \ref{fig:err_vs_K} shows the variation of the misclassification rate of target samples with the number of neighbors ($K$) for different numbers of labeled samples ($N$) in the target domain. The results are averaged over around 10 random repetitions of the experiment with different selections of the labeled samples.

The results in Figure \ref{fig:err_vs_K} show that the SDA-DAGL algorithm performs better than the SDA algorithm in almost all cases. This suggests that even if the SDA-DAGL algorithm is initialized with nonoptimal graphs, it can successfully learn a suitable pair of source and target graphs and accurately estimate the target labels. The performance of SDA-DAGL is seen to be robust to the choice of the initial number of neighbors $K$ in the synthetic data set in Figure \ref{fig:error_vs_K_toy}, whereas it is more affected by the choice of $K$ in the COIL-20 data set in Figure \ref{fig:error_vs_K_coil}. This is because the COIL-20 data set conforms quite well to a low-dimensional manifold structure due to the regular sampling of the camera rotation parameter generating the data set. Initializing the weight matrices with too high $K$ values leads to the loss of the information of the geometric structure from the beginning and makes it more difficult for the algorithm to recover the correct graph topologies along with the label estimates. The fact that too small $K$ values also yield large error in Figure \ref{fig:error_vs_K_coil} can be explained with the incompatibility of small $K$ values with the graph pruning strategy employed in our method. 
We also show in Figure \ref{fig:Wt_iter} the evolution of the weight matrix $W^t$ during two consecutive iterations of the SDA-DAGL method for the COIL-20 data set. The weights of the within-class edges in Class 4 and the between-class edges between Classes 4 and 10 are shown. The update on $W^t$ is seen to mostly preserve the within-class edges, while it removes most of the between-class edges. Thus, the learnt graph topology is progressively improved.


We then study how the difference between the sizes of the source and the target graphs affects the algorithm performance. We fix the number $\Ns$ of source nodes, and vary the number $\Nt$ of target nodes by constructing the target graph with a randomly selected subset of the target samples. The variation of the classification error with $\Nt$ for different $N$ values (number of labeled target nodes) is presented in Table \ref{tab:var_err_Nt}. The results show that the best performance is obtained when the source and the target graphs have an equal number of nodes ($\Nt=\Ns$). In the synthetic data set, the removal of up to $25\%$ of the target graph nodes is seen to be tolerable without much loss in the performance. On the other hand, the performance degrades more severely in the COIL-20 data set as the difference between the source and the target graph sizes increases. Due to the very particular geometric structure of this data set, the algorithm is more sensitive to the dissimilarity between the graph topologies.


\begin{table}[t]
\caption{The variation of the target classification error with the number of target graph nodes. The classification performance tends to be higher when the source and the target graph sizes are similar.}
\begin{minipage}[b]{0.49\linewidth}
\begin{tabular}{l | ccccc}
$\Nt$ \ & 100 & 125 & 150 & 175 & 200  \\ \hline
$N=20$   & 17.13  & 12.67 & 5.31 & 8.51 & 6.67 \\
$N=30$  & 11.57  & 8.52 & 5.50 & 6.62 & 6.17 \\
$N=40$  & 10.17  & 5.53 & 4.82 & 6.81 & 4.31 
\end{tabular}
  \centerline{(a) Synthetic data set, $\Ns=200$}\medskip
\end{minipage}
\hfill
\begin{minipage}[b]{0.49\linewidth}
\begin{tabular}{l | ccccc}
$\Nt$ \ & 30 & 40 & 50 & 60 & 72  \\ \hline
$N=3$   & 28.22 & 25.08 & 22.42 & 17.79 & 6.52 \\
$N=5$  & 27.44  & 24.11 & 15.60 & 14.15 & 1.13 \\
$N=10$  & 22.70  & 18.67 & 16.85 & 5.76 & 1.06 
\end{tabular}
  \centerline{(b) COIL-20 data set, $\Ns=72$ }\medskip
\end{minipage}
\label{tab:var_err_Nt}
\end{table}

\begin{figure}[t]
\begin{center}
     \subfigure[Synthetic data set]
       {\label{fig:Errors_toy}\includegraphics[height=4cm]{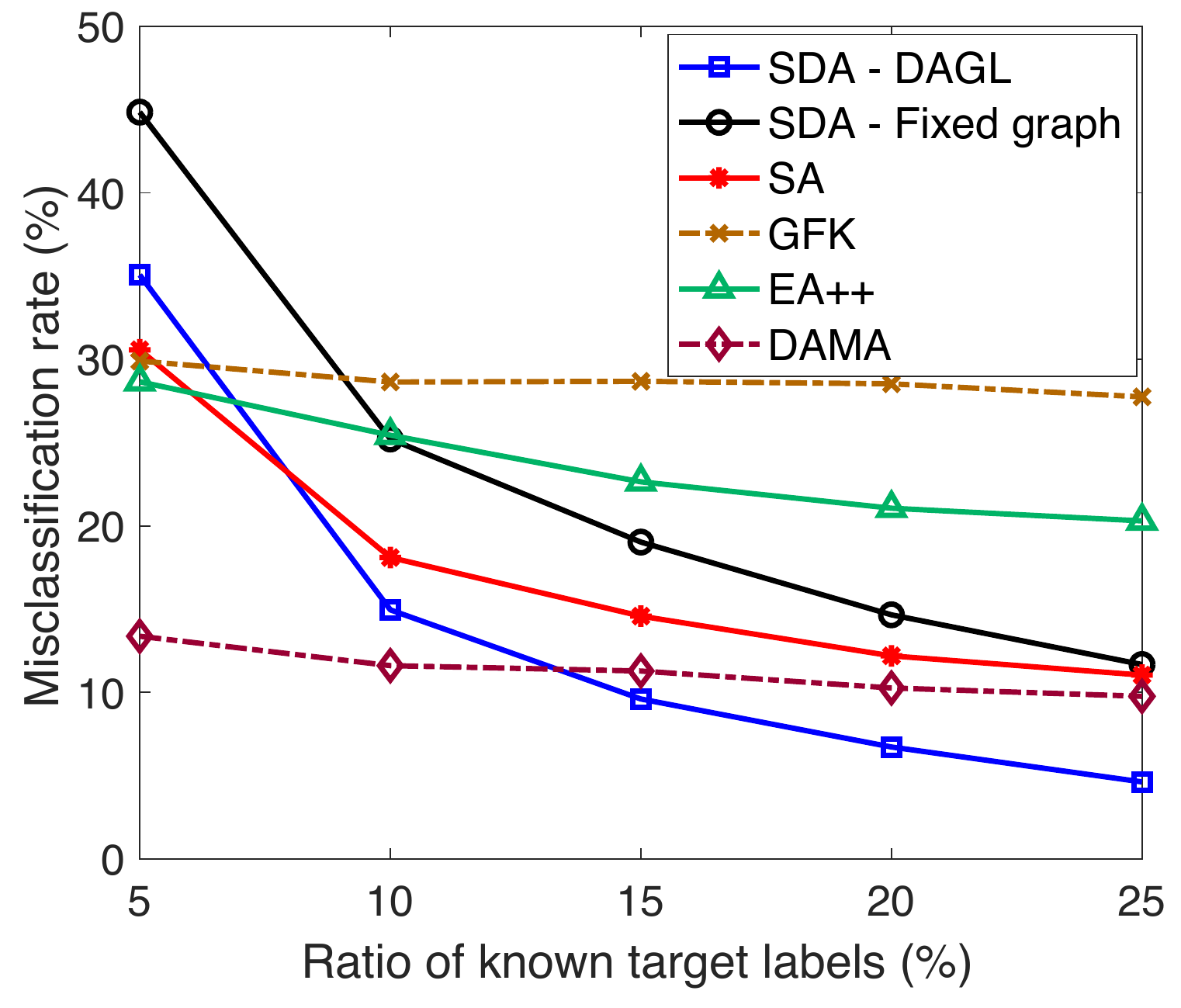}}
      \subfigure[COIL-20 data set]
       {\label{fig:Errors_coil}\includegraphics[height=4cm]{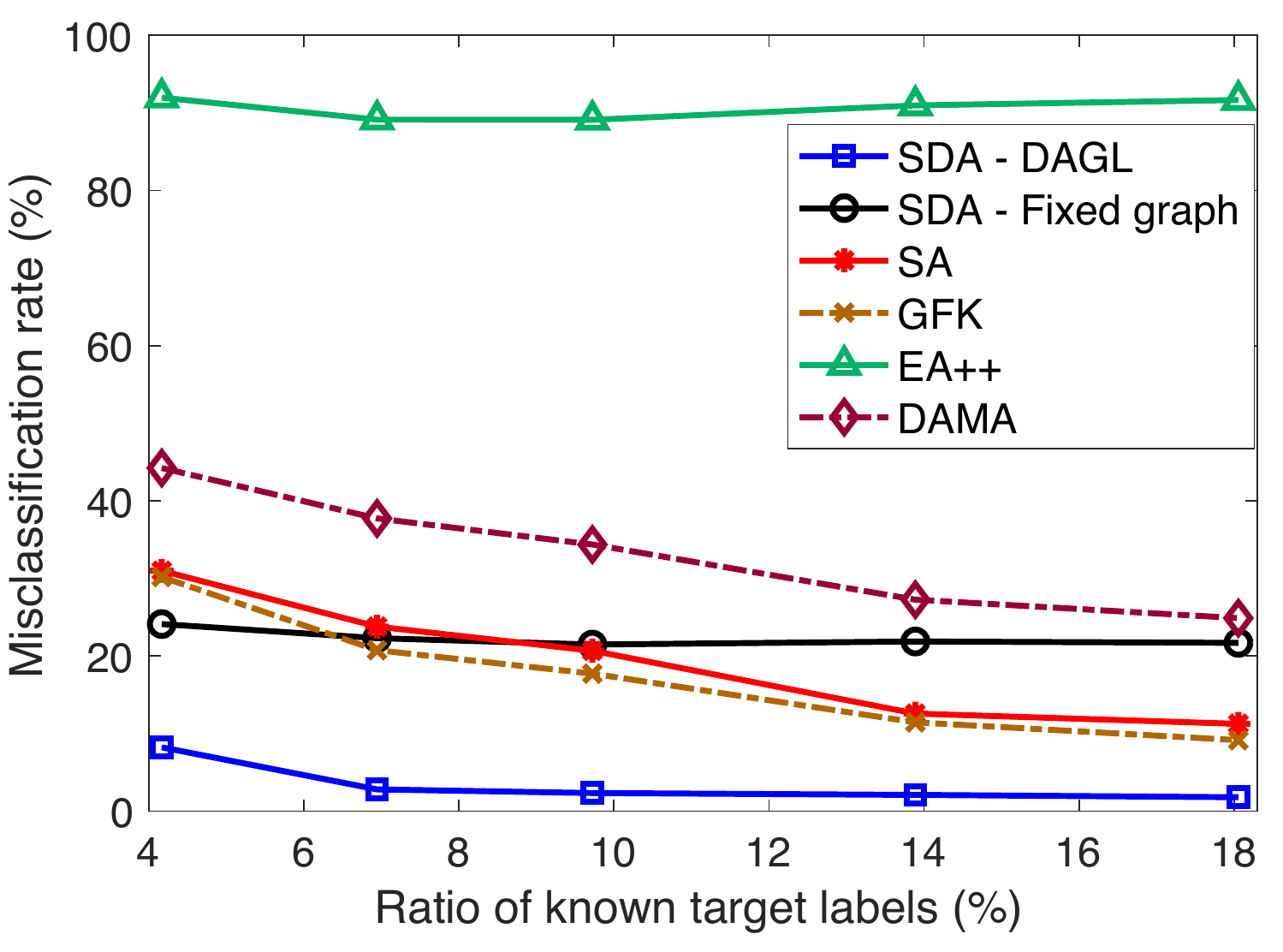}}
 \end{center}
 \vspace{-0.3cm}
 \caption{The variation of the misclassification rate of the target samples with the ratio of labeled target samples. The proposed method is seen to outperform baseline approaches in most cases.}
 \label{fig:err_vs_N}
\end{figure}

We finally present an overall comparison of the proposed SDA-DAGL method with some baseline domain adaptation methods representing different approaches. We compare our method to the Easy Adapt ++ (EA++) \cite{Daume2010} algorithm based on feature augmentation; the Domain Adaptation using Manifold Alignment (DAMA) \cite{WangM11} algorithm which is a graph-based method learning a supervised embedding; and the Geodesic Flow Kernel (GFK) \cite{GongSSG12} and Subspace Alignment (SA) \cite{Fernando2013} methods, which align the PCA bases of the two domains via unsupervised projections.  The misclassification rates of the algorithms on target samples are plotted with respect to the ratio of known target labels in Figures \ref{fig:Errors_toy} and \ref{fig:Errors_coil}, respectively for the synthetic data set and the COIL-20 data sets. The misclassification error decreases as the ratio of the known target labels increases as expected. In both data sets, the proposed SDA-DAGL algorithm is often observed to outperform the baseline approaches and the SDA method, which uses a fixed graph topology. This suggests that the proposed graph learning strategy provides an effective solution for improving the performance of domain adaptation on graphs.

\section{Conclusion}
\label{sec:concl}

In this paper, we have studied the problem of domain adaptation on graphs both theoretically and methodologically. We have first proposed a theoretical analysis of the performance of graph domain adaptation methods. We have considered a setting where a pair of graphs are constructed from data samples drawn from a source manifold and a target manifold. We have focused on a graph domain adaptation framework where the source and the target label functions are estimated such that they have similar spectra. We have proposed an upper bound on the estimation error of the target label function and studied its dependence on the number of data samples, the geometric properties of the data manifolds, and graph parameters such as edge weights and the number of neighbors of graph nodes. In particular, as far as the graph properties are concerned, our theoretical results suggest that a ``balanced'' graph topology improves the performance of learning where the numbers of neighbors are proportionate across different nodes, and too weak edge weights and edges between too distant samples are avoided. Based on these findings, we then proposed a graph domain adaptation algorithm that jointly learns the graph structures and the label functions. Experimental results on synthetic and real data sets suggest that the proposed method yields promising performance in problems concerning machine learning on graph domains. An interesting future direction may be the extension of the study to multiple domains.

\bibliographystyle{IEEEbib}
\bibliography{refs}

\appendix

\section{Proof of Theorem \ref{thm:targetperf}}
\label{pf:thm_targetperf}

In order to prove Theorem \ref{thm:targetperf}, we proceed as follows. We first propose an upper bound on the difference between the rates of variation of the source and target label functions in Section \ref{app:ssec:rates_var_label_fun}. Next, we study the deviation between the eigenvalues of the source and target graph Laplacians in Section \ref{app:ssec:diff_spectra_graphs}. Finally, we  combine these results in Section \ref{app:ssec:bnd_tar_est_err} and present our upper bound on the estimation error of the target label function to finalize the proof.


\subsection{Analysis of the difference between the rates of variation of the source and target label functions}
\label{app:ssec:rates_var_label_fun}

We propose in Lemma \ref{lem:bnd_diff_fsft_var}  an upper bound on the difference between the rates of variation $(\hfs)^T L^s \hfs$ and $(\hft)^T L^t \hft $ of the estimated source and label functions on the graphs. 

\begin{lemma}
\label{lem:bnd_diff_fsft_var}

Let $0= \lambda_1^s \leq \lambda_2^s \leq \dots \leq \lambda_{\sizeU}^s$ and $0 =\lambda_1^t \leq \lambda_2^t \leq \dots \leq \lambda_{\sizeU}^t$ respectively denote the $\sizeU$  smallest  eigenvalues of the source and target graph Laplacians $L^s$ and $L^t$. Assume that the deviation between the corresponding eigenvalues of the two graph Laplacians are bounded as $| \lambda_i^s - \lambda_i^t | \leq \dlambda$, for all $i=1, \dots, \sizeU$. Let the bandwidth parameter $\lambdaR = \max(\lambda_\sizeU^s, \lambda_\sizeU^t)$  indicate an upper bound for the frequencies of the first $\sizeU$ source and target Fourier basis vectors. 

In the estimates $\hfs = \rU^s  \ralpha^s$, $\hft = \rU^t  \ralpha^t$   of the source and target label functions, let  the difference between the Fourier coefficients of the label functions be bounded as $\| \ralpha^s - \ralpha^t \| \leq \dalpha$, and let $\calpha$ be a bound on the norms of the coefficients such that $\| \ralpha^s \|, \| \ralpha^t \| \leq \calpha$. 

Then the difference between the rates of variation of the source and target label function estimates on the graphs can be bounded as  

\begin{equation*}
\begin{split}
|  (\hfs)^T L^s  \hfs - (\hft)^T L^t \hft | \leq  \calpha^2 \dlambda &+ 2 \calpha \lambdaR \dalpha.
\end{split}
\end{equation*}
\end{lemma}

The proof of Lemma \ref{lem:bnd_diff_fsft_var} is an adaptation of the proof of \cite[Proposition 1]{PilanciV18} to our case and it is given in Appendix \ref{pf:lem_bnd_diff_fsft_var}. Lemma \ref{lem:bnd_diff_fsft_var} can be interpreted  as follows: When the source and the target graphs are sufficiently similar, their graph Laplacians will have similar spectra, and the difference between their $i$-th eigenvalues can be bounded as $| \lambda_i^s - \lambda_i^t  | \leq \dlambda$ for a relatively small constant $\dlambda$. If in addition, the label functions vary sufficiently slowly to have limited bandwidth, then the source and target label functions have a similar rate of variation on the two graphs. 


\subsection{Analysis of the difference between the spectra of the source and target graphs}
\label{app:ssec:diff_spectra_graphs}

In Lemma \ref{lem:bnd_diff_fsft_var} we have assumed that the source and target graph Laplacians have similar spectra, so that the difference $| \lambda_i^s - \lambda_i^t  | $ between their eigenvalues  can be suitably bounded. We now determine under which conditions this is possible. We aim to develop an upper bound on $| \lambda_i^s - \lambda_i^t  | $ for all $i$ in terms of the properties of the data manifolds and the constructed graphs.

In domain adaptation problems, a one-to-one correspondence between the source and target data samples is often not available, in contrast to multi-view or multi-modal learning problems. Hence, we do not assume that there exists a particular relation between the samples $x_i^s=\gs(\gamma_{s_i})$ and $x_i^t=\gt(\gamma_{t_i})$, such as being generated from the same parameter vector $\gamma_{s_i}=\gamma_{t_i}$ in the parameter space. Nevertheless, if the manifolds $\Ms$ and $\Mt$ defined over the same parameter space $\Gamma$ are sampled under similar conditions (although independently), one may assume that the independently formed sample sets $\{ \xis \}$ and $\{ \xit \}$ can be reordered so that their corresponding parameter vectors $\{ \gamma_i^s \}$ and $\{ \gamma_i^t \}$ ``fall'' into nearby regions of the parameter space $\Gamma$. We state this assumption with the help of a cover
\[
\mathcal{C} = \bigcup_{m=1}^M B_{\epsilon_m}(\gamma_m) \subset \Gamma
\]
in the parameter space, where $B_{\epsilon}(\gamma)$ denotes an open ball of radius $\epsilon$ around the parameter vector $\gamma$
\[
 B_{\epsilon}(\gamma) = \{ \gamma' \in \Gamma : \| \gamma' - \gamma \| < \epsilon \}
\]
and the parameter vector of any data sample is in at least one ball in $\mathcal{C}$
\[
\{ \gamma_i^s \} \cup \{ \gamma_i^t \} \subset \mathcal{C} = \bigcup_{m=1}^M B_{\epsilon_m}(\gamma_m).
\]
 
We assume without loss of generality that the source and target samples $\{ \xis \}_{i=1}^\N$ and $\{ \xit \}_{i=1}^\N$ are ordered such that for each $i=1, \dots, \N$, 
\[  
\gamma_i^s \in B_{\epsilon_m}(\gamma_m),  \quad \quad \gamma_i^t \in B_{\epsilon_m}(\gamma_m)
\]
the samples $\gamma_i^s $ and  $\gamma_i^t $ are in the same ball  $B_{\epsilon_m}(\gamma_m)$ for some $m \in \{ 1, \dots, M \}$. This condition imposes an ordering of the samples such that source and target samples with nearby indices correspond to nearby parameter vectors in the parameter space $\Gamma$. 

\textbf{Remarks:} 1. First, let us note that when applying a graph-based domain adaptation algorithm, as the parameter-domain representation of the data samples is often unknown, it will not be possible to actually find such a reordering of the data. Nevertheless, our assumption of such a reordering is just for the purpose of theoretical analysis and is not needed in practice. This is because reordering the graph nodes will result in a permutation of the rows and the columns of the graph Laplacian. Since the permutations of the rows and columns with the same indices can be represented as the left and right multiplications of the graph Laplacian with the same symmetric rotation matrix, it will not change its eigenvalues. Hence, our analysis of the deviation $| \lambda_i^s - \lambda_i^t |$ between the eigenvalues under the reordering assumption is still valid even if the nodes are not reordered in practice.    

2. Second, notice that for a finite sampling of the source and target data, it is always possible to find such covers as mentioned above. However, the radii $\epsilon_1, \dots, \epsilon_M$ of the open balls depend on the number of samples. In particular, the radii of these open sets decrease proportionally to the typical inter-sample distance between neighboring manifold samples as the number of samples $\N$ increases, i.e., $\epsilon_m = O(\N^{-1/d})$ where $d$ is the dimension of the manifolds $\Ms, \Mt$.\\

Next, we define some parameters regarding the properties of the graph. Let the source node $\xis$ have $\Kis$ neighbors in the source graph $G^s$, where we denote $\xis \sim \xjs$ when $\xis$ and $\xjs$ are connected with an edge. Similarly let $\Kit$ denote the number of neighbors of $\xit$ in the target graph. Among the $\Kis$ neighbors $\xjs$ of $\xis$ in the source graph, some of their ``correspondences'' $\xjt$ (with respect to the ordering discussed above) will be neighbors of the target node $\xit$ corresponding to $\xis$. Let $\beta_i^s$ be the proportion of such nodes, which is given by
\[
\beta_i^s= \frac{1}{\Kis} \, |  \{ j:  \xjs \sim \xis, \xjt \sim \xit  \}  |
\]
where $| \cdot |$ denotes the cardinality of a set. Similarly, let
\[
\beta_i^t= \frac{1}{\Kit} \, |  \{ j:  \xjt \sim \xit , \xjs \sim \xis \}  |.
\]
The more similar the source and target graphs are, the closer the parameters $\beta_i^s \leq 1$ and $\beta_i^t \leq 1$ will be to $1$.

Finally, we define the parameter $\emax$ as
%
%
\[
\emax= \max \left(  \max_{\xis \sim \xjs} \|  \gamma_i^s - \gamma_j^s \|, 
    \max_{\xit \sim \xjt} \|  \gamma_i^t - \gamma_j^t \|  \right)
\]
%
which gives the largest possible parameter domain distance between two neighboring nodes in the source or target graphs. Also, due to our assumption on the ordering of the samples, for any $i$, there exists some $m \in \{ 1, \dots, M \}$ such that $\gamma_i^s, \gamma_i^t \in B_{\epsilon_{m}}(\gamma_{m})$. Let 
\[
\epsilon_{m_i} = \min \{\epsilon_m : \gamma_i^s, \gamma_i^t \in B_{\epsilon_{m}}(\gamma_{m}) \} 
\]
and also let 
\[
\egen= \emax + 2 \max_i \epsilon_{m_i} 
\]
represent a generic upper bound on the parameter domain distance between ``within-domain'' and ``cross-domain'' neighboring samples. 


We are now ready to state our bound on the deviation $| \lambda_i^s - \lambda_i^t |$ between the corresponding eigenvalues of the source and target graph Laplacians in Lemma \ref{lem:eigval_dev_st}.

\begin{lemma}
\label{lem:eigval_dev_st}
Let $L^s$ and $L^t$ be the Laplacian matrices of the source and target graphs with respective eigenvalues ordered as $0= \lambda_1^s \leq \lambda_2^s \leq \dots \leq \lambda_{\N}^s$, and $0= \lambda_1^t \leq \lambda_2^t \leq \dots \leq \lambda_{\N}^t$. Based on the above definitions of the graph parameters, let us denote
%
\[
\DelW :=   \Lphi \, ( \A  \, \lips  +  \lips + \lipt )   \egen.
\]
Then for all $i=1, \dots, \N$ the deviation $| \lambda_i^s - \lambda_i^t |$ between the corresponding eigenvalues of $L^s$ and $L^t$ is upper bounded as
\[
| \lambda_i^s - \lambda_i^t |
\leq
  \rho_{\max} := \max_{i=1, \dots, \N} \, 2 \left(
 \beta_i^s \Kis  \DelW
 + (1-\beta_i^s) \Kis \phi_0 
 + (1-\beta_i^t) \Kit \phi_0 
 \right).
\]
\end{lemma}

The proof of Lemma \ref{lem:eigval_dev_st} is presented in Appendix \ref{pf:lem_eigval_dev_st}. Lemma \ref{lem:eigval_dev_st} can be interpreted as follows. First, recall that the parameters $\beta_i^s$ and $\beta_i^t$ give the ratio of the source and target neighbors of a graph node that have correspondences in the other graph. Hence, they provide a measure of the similarity between the  source and the target graphs. When the resemblance between the source and target graphs gets stronger, these parameters get closer to $1$. If we consider the asymptotic case where $\beta_i^s$ and $\beta_i^t$ approach $1$, the upper bound $\rho_{\max}$ on the deviation between the eigenvalues becomes 
\[
\lim_{\beta_i^s \rightarrow 1, \,  \beta_i^t \rightarrow 1} \rho_{\max} = \max_i  \, 2 \Kis \DelW.
\]
In order to understand how the eigenvalue deviation changes with the sampling density of the graphs, we can study the variation of this term with the number $\N$ of graph nodes. Assuming that the number of neighbors $\Kis$ are of $O(1)$ with respect to $\N$, i.e., if the number of neighbors is not increased proportionally to $\N$ in the graph construction, the rate of variation of $\rho_{\max}$ with $\N$ is given by that of the term $\DelW$. 
As the number  $\N$ of data points sampled from the manifolds $\Ms$, $\Mt$ increases, we have 
\[
 \epsilon_{m_i},  \emax = O (\N^{-1/d})
\]
and consequently, $\egen = O (\N^{-1/d})$, where $d$ is the intrinsic dimension of the manifolds. This yields $\DelW=O (\N^{-1/d})$, hence, we obtain
\[
| \lambda_i^s - \lambda_i^t |  \leq \rho_{\max} = O (\N^{-1/d})
\]
for the asymptotic case $\beta_i^s \rightarrow 1, \,  \beta_i^t \rightarrow 1$. 

Next, we can also interpret Lemma \ref{lem:eigval_dev_st} from the perspective of the similarity between the manifolds. As the geometric structures of the source and target manifolds $\Ms$, $\Mt$ get more similar, the constants $\Al$ and $\Au$ will approach $1$, and the constant $A$ will approach 0. We observe that this reduces the deviation $| \lambda_i^s - \lambda_i^t |$ as $\DelW$ is proportional to $A$. Also, as the Lipschitz regularity of the functions $\gs$, $\gt$ defining the source and target manifolds improves, the constants $\lips$ and $\lipt$ will decrease, hence, $\DelW$ will also decrease. We can summarize these with the observation that as $\lips, \lipt \rightarrow 0$, $\A \rightarrow 0$, and $\beta_i^s, \beta_i^t \rightarrow 1$, the eigenvalue difference  converges to 0 as $| \lambda_i^s - \lambda_i^t | \rightarrow 0$ for all $i=1, \dots, N.$

\subsection{Bounding the estimation error of the target label function}
\label{app:ssec:bnd_tar_est_err}

Putting together the results from Lemmas \ref{lem:bnd_diff_fsft_var} and \ref{lem:eigval_dev_st}, we arrive at the following observation: Assuming that the conditions of Lemma  \ref{lem:eigval_dev_st} hold, the spectrum perturbation parameter $\dlambda$ in Lemma \ref{lem:bnd_diff_fsft_var} can be upper bounded as $\dlambda \leq   \rho_{\max} $. In this case, the rates of variations of the source and target function estimates differ as
\[
|  (\hfs)^T L^s  \hfs - (\hft)^T L^t \hft | \leq  \calpha^2 \rho_{\max} + 2 \calpha \lambdaR \dalpha.
\]
Due to our assumption $\fs=\hfs$, the above inequality implies
\[
(\hft)^T L^t \hft \leq  \hB = (\fs)^T L^s  \fs  +  \calpha^2 \rho_{\max} + 2 \calpha \lambdaR \dalpha.
\]
The parameter $\hB$ thus gives a bound on the rate of variation of the label function estimate $\hft$ on the graph. 
%
%
Recall also the assumption on the true target label function in Theorem \ref{thm:targetperf}
\[
(\ft)^T L^t \ft \leq  \B.
\]
Under these assumptions, we would like to find an upper bound on the target label estimation error $\| \hft - \ft \|$. 

Before we state our bound on the estimation error, we first define some parameters related to the properties of the target graph. First, recall from Section \ref{ssec:notation} that we consider a setting where very few labels $y^t_i = f^t(x^t_i) $ are known on the target graph, with indices $i \in \labT \subset \{ 1, \dots, \Nt \}$ in the index set $\labT $.  Now let us denote the set of indices of the unlabeled nodes of the target graph as 
$
\ulabT =  \indT \setminus \labT,
$
where $\indT = \{ 1, \dots, \Nt \} $ denotes the set of all target node indices and $\cdot \setminus \cdot$ denotes the set difference. Moreover, assuming that the target graph is connected, or otherwise each connected component contains at least one labeled node, one can partition the node indices as
$
\indT = \labT \cup \ulaboneT \cup \dots \cup \ulabQT,
$
where $\ulabqT$ consists of the indices of unlabeled nodes that are connected to the nearest labeled node through a shortest path of length $\q$ (with $\q$ hops). $\Q$ is then the longest possible length of the shortest path between an unlabeled and a labeled node. Let $\ulabzerT := \labT$ simply stand for the index set of labeled nodes. We thus partition the target graph nodes in sets of several ``layers'' $\q=0, 1, \dots, \Q$, with respect to their proximity to the closest labeled node.

Let
$
\Neig_j^\q := \{  i \in  \ulabqT :  \xit \sim \xjt \}
$
denote the set of indices of the neighbors of a node $\xjt$ within the nodes of layer $\q$, where $\sim$ denotes neighboring nodes. We can define the parameters
$
\Kminq := \min_{j \in \ulabqT} | \Neig_j^{\q-1} |
$
and
$
\Kmaxq := \max_{j \in \ulabqT} | \Neig_j^{\q+1} |, 
$
where $| \cdot |$ denotes the cardinality. $\Kminq$ is thus the minimum number of neighbors between a layer $\q$ and its preceding layer $\q-1$, and $\Kmaxq$ is the maximum number of neighbors between a layer $\q$ and its succeeding layer $\q+1$.
Let us also define the minimum weight of an edge between any two nodes from consecutive layers  as
$
\wmin := \min_{\q = 1, \dots, \Q} \wmin_\q,
$
where 
$
\wmin_\q =  \min \{ W^t_{ij} : \xit \sim \xjt, \, i \in \ulabqT, \,   j \in \ulabqmoT  \}
$
is the smallest edge weight between layers $\q$ and $\q-1$.

Our derivation of an upper bound on the target label estimation error is based on the following decomposition
\[
\| \hft - \ft \|^2 = \sum_{\q = 0}^\Q  \| \hft_{U_\q} - \ft_{U_\q} \|^2
\]
where the vectors $\ft_{U_\q}$ and $\hft_{U_\q}$ are respectively obtained by restricting the vectors $\ft$ and $\hft$ to their entries in the index set $\ulabqT$. Our strategy for bounding the target error is to bound the error of each layer in terms of the error of the previous layer, and use the observation that the error of layer $q=0$ is $0$ as it consists of the labeled samples. In the following result, we first provide an upper bound on the error $\| \hft_{U_\q} - \ft_{U_\q} \|^2$ of the $\q$-th layer in terms of the error $\| \hft_{U_{\q-1}} - \ft_{U_{\q-1}}  \|^2$ of the preceding layer $\q-1$.

\begin{lemma}
\label{lem:targ_err_ito_prev}
Let 
\[
\B_\q := \sum_{j \in \ulabqT}  \  \sum_{ \xit \sim \xjt, \ i \in \ulabqmoT } W_{ij}^t (\ft_i - \ft_j)^2
\]
denote the total rate of variation of the true target label function $\ft$ over the edges between two consecutive layers $\q$ and $\q-1$. Similarly define 
\[
\hB_\q := \sum_{j \in \ulabqT}  \  \sum_{ \xit \sim \xjt, \ i \in \ulabqmoT } W_{ij}^t (\hft_i - \hft_j)^2
\]
for the estimate $\hft$ of the target label function. We can then bound the estimation error of the $\q$-th layer in terms of the estimation error of the preceding layer $\q-1$ as
\[
\|  \hft_{U_\q} - \ft_{U_\q}  \|^2   \leq   
 \frac{| \ulabqT |}{\Kminq}
 \left (  
 \frac{ \sqrt{\B_\q}  + \sqrt{\hB_\q} }{\sqrt{ \wmin_\q} } 
+ \sqrt{ \Kmaxqmo }  \ \|  \hft_{U_{\q-1}} - \ft_{U_{\q-1}} \|
\right )^2
\]
for $\q=1, \dots, \Q$.
\end{lemma}

The proof of Lemma \ref{lem:targ_err_ito_prev} is given in Appendix \ref{pf:lem_targ_err_ito_prev}.

Lemma \ref{lem:targ_err_ito_prev} provides an upper bound on the error of each layer in terms of the error of the previous layer. We can now use this result to obtain an upper bound on the overall target label estimation error, which is presented in the following lemma.

\begin{lemma}
\label{lem:targ_est_err}

The estimation error of the target label function can be bounded as
\[
\| \ft  - \hft \|^2 \leq 
\frac{\kappa}{\wmin} (\sqrt{\B} + \sqrt{\hB})^2
\]
where
\[
\kappa = \sum_{\q=1}^\Q 
\frac{| \ulabqT |}{\Kminq} 
 \left( 1
 +  \sum_{l=1}^{\q-1}       \prod_{m=l}^{\q-1}  |  I_{U_m}^t   |   \frac{K_m^{\max}}{K_m^{\min}} 
 \right) .
\]
\end{lemma}

The proof of Lemma \ref{lem:targ_est_err} is given in Appendix \ref{pf:lem_targ_est_err}. 

The results stated in Lemmas \ref{lem:bnd_diff_fsft_var}-\ref{lem:targ_est_err} provide a complete characterization of the performance of estimating the target label function in a graph domain adaptation setting. We are now ready to combine these results in the following  main result.

\begin{theorem}
\label{thm:targetperf_long}

Consider a graph-based domain adaptation algorithm matching the spectra of the source and target label functions $\hfs= \rU^s \ralpha^s$ and $\hft= \rU^t \ralpha^t$ such that the difference between the Fourier coefficients of the label functions are bounded as $\| \ralpha^s - \ralpha^t \| \leq \dalpha$, the norms of the Fourier coefficients are bounded as $\| \ralpha^s \|, \| \ralpha^t \| \leq \calpha$, and $\hfs$ and $\hft$ are band-limited on the graphs so as not to contain any components with frequencies larger than $\lambdaR$.

Assume the source and target graphs are constructed independently from equally many data samples by setting the graph weights via the kernel $\phi$, where the source samples $\{  \xis \}_{i=1}^\N$ and target samples $\{  \xit \}_{i=1}^\N$ are drawn from the manifolds $\Ms$ and $\Mt$ defined via the functions $\gs$ and $\gt$ over a common parameter domain $\Gamma$. Let
\begin{equation}
\DelW =   \Lphi \, ( \A  \, \lips  +  \lips + \lipt )  \, \egen.
\end{equation}
and
\begin{equation}
  \rho_{\max} = \max_{i=1, \dots, \N} \, 2 \left(
 \beta_i^s \Kis  \DelW
 + (1-\beta_i^s) \Kis \phi_0 
 + (1-\beta_i^t) \Kit \phi_0 
 \right).
\end{equation}
Then, the difference between the rates of variation of the source and target function estimates is upper bounded as
\[
|  (\hfs)^T L^s  \hfs - (\hft)^T L^t \hft | \leq  \calpha^2 \rho_{\max} + 2 \calpha \lambdaR \dalpha.
\]
Moreover, if $\hfs=\fs$ and if the target label function estimates $\hft_i$ are equal to the true labels $\ft_i$ at labeled nodes, the target label estimation error can be bounded as
\begin{equation}
\| \hft  - \ft \|^2 \leq 
\frac{\kappa}{\wmin} (\sqrt{\B} + \sqrt{\hB})^2
\end{equation}
where
\begin{equation}
\hB = (\fs)^T L^s  \fs  +  \calpha^2 \rho_{\max} + 2 \calpha \lambdaR \dalpha,
\end{equation}
$\B$ is an upper bound on the rate of variation of the true label function
\begin{equation}
(\ft)^T L^t \ft \leq  \B,
\end{equation}
and
\[
\kappa = \sum_{\q=1}^\Q 
\frac{| \ulabqT |}{\Kminq} 
 \left( 1
 +  \sum_{l=1}^{\q-1}       \prod_{m=l}^{\q-1}  |  I_{U_m}^t   |   \frac{K_m^{\max}}{K_m^{\min}} 
 \right) .
\]
\end{theorem}

We finally conclude the proof of Theorem \ref{thm:targetperf} by observing that it is simply a summarizing restatement of the result in Theorem \ref{thm:targetperf_long}.

\section{Proofs of the Lemmas in Appendix \ref{pf:thm_targetperf}} 

\subsection{Proof of Lemma \ref{lem:bnd_diff_fsft_var}}
\label{pf:lem_bnd_diff_fsft_var}

\begin{proof}
The rates of variation of $\hfs$ and $\hft$ on the source and target graphs are given by
\begin{equation*}
\begin{split}
(\hfs)^T L^s \hfs &=   (\ralpha^s)^T  (\rU^s)^T  L^s \rU^s  \ralpha^s =  (\ralpha^s)^T  \rLambda^s  \ralpha^s \\
(\hft)^T L^t \hft &=   ( \ralpha^t)^T    (\rU^t)^T  L^t \rU^t  \ralpha^t =  (\ralpha^t)^T  \rLambda^t   \ralpha^t
\end{split}
\end{equation*}
where $\rLambda^s$ and $\rLambda^t$ are the diagonal matrices consisting of the $\sizeU$ smallest eigenvalues of respectively $L^s$ and $L^t$, such that $\rLambda^s_{ii} = \lambda_i^s$ and $\rLambda^t_{ii} = \lambda_i^t$, for $i=1, \dots, \sizeU$.

The difference between the rates of variations of $\hfs$ and $\hft$ can then be bounded as
\begin{equation}
\label{eq:fs_ft_sep3terms}
\begin{split}
& | (\hfs)^T L^s \hfs  -  (\hft)^T L^t \hft   | = |  (\ralpha^s)^T  \rLambda^s  \ralpha^s -   ( \ralpha^t)^T   \rLambda^t   \ralpha^t  | \\
&= | 
  (\ralpha^s)^T  \rLambda^s  \ralpha^s 
 -  (\ralpha^s)^T  \rLambda^t  \ralpha^s  +   (\ralpha^s)^T  \rLambda^t  \ralpha^s    
 -   (\ralpha^t)^T  \rLambda^t  \ralpha^t   
 | \\
& \leq  
| (\ralpha^s)^T ( \rLambda^s -  \rLambda^t) \ralpha^s  |
+ |  (\ralpha^s)^T \rLambda^t \ralpha^s -  (\ralpha^t)^T \rLambda^t \ralpha^t   |.
\end{split}
\end{equation}

In the following, we derive an upper bound for each one of the two terms at the right hand side of the inequality in \eqref{eq:fs_ft_sep3terms}. The first term is bounded as
\begin{equation*}
|  (\ralpha^s)^T ( \rLambda^s -  \rLambda^t) \ralpha^s   | 
\leq \| \ralpha^s \|^2  \| \rLambda^s -  \rLambda^t  \|
\leq  \calpha^2 \dlambda.
\end{equation*}
Here the first inequality is due to the Cauchy-Schwarz inequality, and the second inequality follows from the fact that the operator norm of the diagonal matrix $\rLambda^s -  \rLambda^t $ is given by the magnitude of its largest eigenvalue, which cannot exceed $\dlambda$ due to the assumption $| \lambda_i^s - \lambda_i^t | \leq \dlambda$ for all $i$.

Next, we bound the second term in \eqref{eq:fs_ft_sep3terms} as 
\begin{equation*}
\begin{split}
 |  (\ralpha^s)^T \rLambda^t \ralpha^s -  (\ralpha^t)^T \rLambda^t \ralpha^t    | 
&= |  
(\ralpha^s)^T \rLambda^t \ralpha^s  - (\ralpha^s)^T \rLambda^t \ralpha^t
+
(\ralpha^s)^T \rLambda^t \ralpha^t  - (\ralpha^t)^T \rLambda^t \ralpha^t
| \\
& \leq 
|  (\ralpha^s)^T \rLambda^t  (\ralpha^s -  \ralpha^t) |
+ | (\ralpha^s -  \ralpha^t)^T  \rLambda^t  \ralpha^t  | \\
& \leq \|  \ralpha^s \|   \| \rLambda^t \|     \|  \ralpha^s -  \ralpha^t  \|  
+  \|  \ralpha^s -  \ralpha^t  \|   \| \rLambda^t \|   \|  \ralpha^t \| 
\leq 2 \calpha \lambdaR \dalpha
\end{split}
\end{equation*}
where the last equality follows from the fact that the matrix norm $\|  \rLambda^t \|$ is given by the largest eigenvalue of $\rLambda^t $, which is smaller than $\lambdaR$ by our assumption.

Putting together the upper bounds for both terms in \eqref{eq:fs_ft_sep3terms}, we get the stated result
\begin{equation*}
\begin{split}
 | (\hfs)^T L^s \hfs  -  (\hft)^T L^t \hft   |  \leq 
 \calpha^2 \dlambda + 2 \calpha \lambdaR \dalpha  .
\end{split}
\end{equation*}

\end{proof}

\subsection{Proof of Lemma \ref{lem:eigval_dev_st}}
\label{pf:lem_eigval_dev_st}

\begin{proof}
In order to show that the stated upper bound holds on the difference between the eigenvalues, we first examine the difference $|W^s_{ij} - W^t_{ij}|$ between the corresponding entries of the source and target weight matrices. We propose an upper bound on $|W^s_{ij} - W^t_{ij}|$ for three different cases below where at least one of $W^s_{ij}$ and  $W^t_{ij}$ is nonzero.

\textit{Case 1.} When the source samples $\xis \sim \xis$ are neighbors on the source graph and the corresponding target samples  $\xit \sim \xjt$ are also neighbors on the target graph at the same time, we bound $|W^s_{ij} - W^t_{ij}|$ as
\begin{equation}
\label{eq:form_wsij_wtij}
\begin{split}
| W^s_{ij} - W^t_{ij}  | &= \left | \phi(\| \xis - \xjs \|) - \phi( \| \xit - \xjt   \| )  \right | 
\leq \Lphi \left | \| \xis - \xjs \|  - \| \xit - \xjt \|  \right |   .\\
\end{split}
\end{equation}

We proceed by examining each one of the terms $\| \xis - \xjs \|$ and  $\| \xit - \xjt \|$. We have
\begin{equation*}
\begin{split}
\| \xis - \xjs \| &= \|  \gs(\gamma_i^s) -   \gs(\gamma_j^s)   \|  \\
&= \|  \gs(\gamma_i^s) - \gs(\gamma_{m_i}) +  \gs(\gamma_{m_i}) 
  -  \gs(\gamma_{m_j}) +  \gs(\gamma_{m_j}) -  \gs(\gamma_j^s)  \|
\end{split}
\end{equation*}
which implies
\begin{equation}
\label{eq:bnd_diff_xijs}
\begin{split}
 \| \gs(\gamma_{m_i})  -  \gs(\gamma_{m_j}) \| -  \Delta_i^s -  \Delta_j^s 
& \leq \| \xis - \xjs \|   \\
& \leq \| \gs(\gamma_{m_i})  -  \gs(\gamma_{m_j}) \|  + \Delta_i^s +  \Delta_j^s
\end{split}
\end{equation}
where
\[
 \Delta_i^s := \|  \gs(\gamma_i^s) - \gs(\gamma_{m_i})  \|,
 \qquad \qquad
 \Delta_j^s := \|     \gs(\gamma_j^s) - \gs(\gamma_{m_j})  \| .
 \]
With a similar derivation, we get
\begin{equation}
\label{eq:bnd_diff_xijt}
\begin{split}
 \| \gt(\gamma_{m_i})  -  \gt(\gamma_{m_j}) \| -  \Delta_i^t -  \Delta_j^t 
& \leq \| \xit - \xjt \|   \\
& \leq \| \gt(\gamma_{m_i})  -  \gt(\gamma_{m_j}) \|  + \Delta_i^t +  \Delta_j^t
\end{split}
\end{equation}
where
\[
 \Delta_i^t := \|  \gt(\gamma_i^t) - \gt(\gamma_{m_i})  \|,
 \qquad \qquad
 \Delta_j^t := \|     \gt(\gamma_j^t) - \gt(\gamma_{m_j})  \| .
 \]
From \eqref{eq:bnd_diff_xijs} and \eqref{eq:bnd_diff_xijt}, we get

\begin{equation}
\label{eq:diff_deltaxijs_xijt}
\begin{split}
\left | \| \xis - \xjs \|  - \| \xit - \xjt \|  \right |   
&\leq
\left |  \| \gs(\gamma_{m_i})  -  \gs(\gamma_{m_j}) \|   -   \| \gt(\gamma_{m_i})  -  \gt(\gamma_{m_j}) \|    \right |  \\
&+  \Delta_i^s +   \Delta_j^s +  \Delta_i^t + \Delta_j^t .
\end{split}
\end{equation}
From the definition \eqref{defn:al_au} of $\Al$ and $\Au$, we have
\[
\Al \| \gs(\gamma_{m_i})  -  \gs(\gamma_{m_j}) \| 
  \leq  \| \gt(\gamma_{m_i})  -  \gt(\gamma_{m_j}) \|  \leq
\Au \| \gs(\gamma_{m_i})  -  \gs(\gamma_{m_j}) \| 
\]
which yields the following bound on the first term of the right hand side of the inequality in \eqref{eq:diff_deltaxijs_xijt}:
\begin{equation}
\label{eq:bnd_gsdiff_gtdiff}
\begin{split}
&\left |  \| \gs(\gamma_{m_i})  -  \gs(\gamma_{m_j}) \|   -   \| \gt(\gamma_{m_i})  -  \gt(\gamma_{m_j}) \|    \right |  \\
&\leq \max(| 1-\Al |, |\Au -1 | ) \,  \| \gs(\gamma_{m_i})  -  \gs(\gamma_{m_j}) \|  \\
&= \A \, \| \gs(\gamma_{m_i})  -  \gs(\gamma_{m_j}) \|  
\leq \A  \, \lips \|  \gamma_{m_i} -  \gamma_{m_j} \| \\
& \leq \A  \, \lips \left (   \|  \gamma_{m_i} -  \gamma_i^s \| +   \|  \gamma_i^s -  \gamma_j^s \|   
 +  \|  \gamma_j^s -  \gamma_{m_j} \|        \right) \\
& \leq  \A  \, \lips ( \epsilon_{m_i} + \emax + \epsilon_{m_j} ).
\end{split}
\end{equation}
The other terms in \eqref{eq:diff_deltaxijs_xijt} can be bounded as
\[
 \Delta_i^s = \|  \gs(\gamma_i^s) - \gs(\gamma_{m_i})  \| 
 \leq \lips \|  \gamma_i^s - \gamma_{m_i} \|
 \leq \lips \, \epsilon_{m_i}
\]
since $\gamma_i^s \in B_{ \epsilon_{m_i}} (\gamma_{m_i})$. Similarly, 
\begin{eqnarray*}
 \Delta_j^s \leq \lips \,  \epsilon_{m_j} ,
 \qquad
 \Delta_i^t \leq \lipt \,  \epsilon_{m_i} 
 \qquad
\Delta_j^t \leq \lipt \,  \epsilon_{m_j}.
\end{eqnarray*}
Using these bounds in \eqref{eq:diff_deltaxijs_xijt} together with the bound in \eqref{eq:bnd_gsdiff_gtdiff}, we get
\begin{equation*}
\begin{split}
\left | \| \xis - \xjs \|  - \| \xit - \xjt \|  \right |   \leq
\A  \, \lips ( \epsilon_{m_i} + \emax + \epsilon_{m_j} )
+  (\lips + \lipt) \, (\epsilon_{m_i} + \epsilon_{m_j})
\end{split}
\end{equation*}
which gives the following bound on $| W^s_{ij} - W^t_{ij}  |$ from \eqref{eq:form_wsij_wtij}
\begin{equation}
\begin{split}
| W^s_{ij} - W^t_{ij}  | &\leq \Lphi \, \A  \, \lips ( \epsilon_{m_i} + \emax + \epsilon_{m_j} )
+  \Lphi (\lips + \lipt) \, (\epsilon_{m_i} + \epsilon_{m_j}) \\
& \leq  \Lphi \, \A  \, \lips \egen +  \Lphi (\lips + \lipt) \egen
= \DelW.
\end{split}
\end{equation}

\textit{Case 2.} When $\xis \sim \xjs$ are neighbors on the source graph but $\xit \nsim \xjt$ are not neighbors on the target graph, $W_{ij}^t =0$, and hence we have
\[
| W_{ij}^s - W_{ij}^t  | = | W_{ij}^s | = \phi ( \| \xis - \xjs \|) \leq \phi(0)= \phi_0.
\]

\textit{Case 3.} Similarly to Case 2, when $\xit \sim \xjt$ are neighbors on the target graph but $\xis \nsim \xjs$ are not neighbors on the source graph, it is easy to obtain
\[
| W_{ij}^s - W_{ij}^t  | \leq  \phi_0.
\]

Having examined all three cases, we can now derive a bound on the difference $| \lambda_i^s - \lambda_j^s |$ between the corresponding source and target Laplacian eigenvalues. Defining
\[
 P = L^t - L^s 
\]
we can write $L^t = L^s +P$, where the difference $P$ can be seen as a ``perturbation'' on the source Laplacian matrix $L^s$. The spectral radius (the largest eigenvalue magnitude) of $P$ can be bounded as
\[
\rho(P) \leq \max_i  \sum_j | P_{ij}  |.
\]
The diagonal entries of the perturbation matrix are given by
\begin{eqnarray*}
P_{ii} &=& L^t_{ii} - L^s_{ii} = D^t_{ii} - D^s_{ii} = \sum_{j \neq i} ( W^t_{ij} - W^s_{ij} ) 
\end{eqnarray*}
whereas the off-diagonal entries are given by
\[
P_{ij}=L^t_{ij} - L^s_{ij} = W^s_{ij}  - W^t_{ij}
\]
for $j \neq i$. Then, for $i=1, \dots, \N$, we have
\begin{equation}
\begin{split}
& \sum_j | P_{ij}  | =  | P_{ii} | + \sum_{j \neq i} | P_{ij}  | \\
  	&=  \left |    \sum_{j \neq i} ( W^t_{ij} - W^s_{ij} )  \right |  +  \sum_{j \neq i} | W^s_{ij}  - W^t_{ij}  | 
	 \leq 2 \sum_{j \neq i} | W^s_{ij}  - W^t_{ij}  |  \\
	&= 2 \left ( \sum_{\xis \sim \xjs, \xit \sim \xjt} | W^s_{ij}  - W^t_{ij}  |  
	+  \sum_{\xis \sim \xjs, \xit \nsim \xjt} | W^s_{ij}  - W^t_{ij}  |  
	+\sum_{\xis \nsim \xjs, \xit \sim \xjt} | W^s_{ij}  - W^t_{ij}  |  
	 \right).
\end{split}
\end{equation}
Using the bounds on the term $ | W^s_{ij}  - W^t_{ij}  |  $ for each one of the three studied cases in the above expression, we get
\begin{equation}
\begin{split}
 \sum_j | P_{ij}  | \leq 2 \left(
 \beta_i^s \Kis  \DelW
 + (1-\beta_i^s) \Kis \phi_0 
 + (1-\beta_i^t) \Kit \phi_0 
 \right)
 \leq  \rho_{\max}
\end{split}
\end{equation}
where the last inequality follows from the definition of $ \rho_{\max}$ in Lemma \ref{lem:eigval_dev_st}. We can then bound the spectral radius of the perturbation matrix as
\[
\rho(P) \leq \max_i  \sum_j | P_{ij}  | \leq \rho_{\max}.
\]
Finally, from Weyl's inequality, the difference between the corresponding eigenvalues $\lambda_i^s$ and $\lambda_i^t$ of the matrices $L^s$ and $L^t$ are upper bounded by the spectral radius of the perturbation matrix. Hence, we have
\[
| \lambda_i^s - \lambda_i^t | \leq \rho(P)  \leq \rho_{\max}
\]
which proves the stated result.

\end{proof}


\subsection{Proof of Lemma \ref{lem:targ_err_ito_prev}}
\label{pf:lem_targ_err_ito_prev}

\begin{proof}
Let us first define the following parameter
\[
\af_j^\q :=  \left( \sum_{\xit \sim \xjt, \ i \in \ulabqT} (\ft_i - \ft_j)^2 \right)^{1/2}
\]
which gives the total difference of the target label function between a node $\xjt$ and its neighbors from the $\q$-th layer. Similarly, let
\[
\haf_j^\q :=  \left( \sum_{\xit \sim \xjt, \ i \in \ulabqT} (\hft_i - \hft_j)^2 \right)^{1/2}
\]
for the estimate of the target label function. Let us also define vectors $\Af^\q \in \R^{|\ulabqT|}$ and  $\hAf^\q \in \R^{|\ulabqT|}$, respectively consisting of the values $\af_j^{\q-1} $ and $\haf_j^{\q-1}$ in their entries, where $j$ varies in the index set $\ulabqT$. We can then lower bound the parameter $\B_\q$ as
\begin{equation}
\begin{split}
\B_\q &= \sum_{j \in \ulabqT}  \  \sum_{ \xit \sim \xjt, \ i \in \ulabqmoT } W_{ij}^t (\ft_i - \ft_j)^2 \\
  &\geq \sum_{j \in \ulabqT}  \  \sum_{ \xit \sim \xjt, \ i \in \ulabqmoT } \wmin_\q (\ft_i - \ft_j)^2 \\
  &\geq \sum_{j \in \ulabqT}  \wmin_\q  \  (\af_j^{\q-1} )^2 
  = \wmin_\q \, \| \Af^\q   \|^2
\end{split}
\end{equation}
which gives
\begin{equation*}
\| \Af^\q   \| \leq \left(  \frac{\B_\q}{ \wmin_\q}   \right)^{1/2}.
\end{equation*}
With similar derivations, we also get
\begin{equation*}
\| \hAf^\q   \| \leq \left(  \frac{\hB_\q}{ \wmin_\q}   \right)^{1/2}.
\end{equation*}
Now, let $\ft_{\Neig_j^\q }$ and $\hft_{\Neig_j^\q }$ respectively denote the restrictions of the vectors $\ft$ and $\hft$ to the indices in $\Neig_j^\q$. Let us fix a node index $j \in \ulabqT $ in the $\q$-th layer. Defining $\ones$ to be a vector of appropriate size consisting of 1's in all its entries,  we obtain the following relation by applying the triangle inequality
\begin{equation}
\label{eq:fj_error_vect}
\begin{split}
\|  \ft_j \ones - \hft_j \ones \|  
&\leq 
\| \ft_j \ones - \ft_{\Neig_j^{\q-1} } \|  + 
\| \ft_{\Neig_j^{\q-1} }   -  \hft_{\Neig_j^{\q-1} }   \|  + 
\| \hft_{\Neig_j^{\q-1} }   - \hft_j \ones   \| \\
&= \af_j^{\q-1} +  \haf_j^{\q-1} + e_j^{\q-1}
\end{split}
\end{equation}
where the equality follows from the definitions of the parameters $\af_j^{\q}$, $\haf_j^{\q}$; and the definition
\begin{equation*}
e_j^{\q-1} := \| \ft_{\Neig_j^{\q-1} }   -  \hft_{\Neig_j^{\q-1} }   \| 
\end{equation*}
of the total estimation error at the neighbors of node $\xjt$ at layer $\q-1$. Observing that the constant vector at the left hand side of the inequality in \eqref{eq:fj_error_vect} consists of $| \Neig_j^{\q-1} |$ entries, we get
\begin{equation*}
|  \ft_j  - \hft_j  | \leq \frac{\af_j^{\q-1} +  \haf_j^{\q-1} + e_j^{\q-1}}{ \sqrt{ | \Neig_j^{\q-1}| }}.
\end{equation*}

We can then bound the estimation error of the $\q$-th layer as 
\begin{equation}
\label{eq:fuq_error_v1}
\begin{split}
\|  \hft_{U_\q} - \ft_{U_\q}  \|^2  &= \sum_{j \in \ulabqT  } (\ft_j  - \hft_j )^2
\leq  \sum_{j \in \ulabqT  }   \frac{ ( \af_j^{\q-1} +  \haf_j^{\q-1} + e_j^{\q-1} )^2 }{  | \Neig_j^{\q-1}| } \\
 & \leq \left( \sum_{j \in \ulabqT  } ( \af_j^{\q-1} +  \haf_j^{\q-1} + e_j^{\q-1} )^2  \right)
 \left( \sum_{j \in \ulabqT  } \frac{1}{| \Neig_j^{\q-1}|}  \right)
\end{split}
\end{equation}
We proceed by upper bounding each one of the terms at the right hand side of the above inequality. In order to bound the first term, let us first define the vector $E^\q \in  \R^{|\ulabqT|}$, which is made up of the entries $e_j^{\q-1}$, where $j$ varies in the set $\ulabqT$. We then have
\begin{equation}
\label{eq:aq_haq_eq_terms}
\begin{split}
& \sum_{j \in \ulabqT  } ( \af_j^{\q-1} +  \haf_j^{\q-1} + e_j^{\q-1} )^2 
= \| \Af^\q + \hAf^\q + E^\q  \|^2 \\
& \leq  ( \| \Af^\q \|  + \|  \hAf^\q   \|   + \| E^\q   \|  )^2
\leq  \left (     
\sqrt{  \frac{\B_\q}{ \wmin_\q} }  + \sqrt{ \frac{\hB_\q}{ \wmin_\q} } + \| E^\q   \|
\right )^2.
\end{split}
\end{equation}
We can relate the term $\| E^\q   \|$ to the estimation error of the preceding layer as
\begin{equation}
\begin{split}
 \| E^\q   \|^2 
 &= \sum_{j \in \ulabqT  } (e_j^{\q-1})^2
   =  \sum_{j \in \ulabqT  }   \| \ft_{\Neig_j^{\q-1} }   -  \hft_{\Neig_j^{\q-1} }   \|^2 \\
 & =  \sum_{j \in \ulabqT  } \ \sum_{  \xit \sim \xjt, \ i \in \ulabqmoT   } (\ft_i - \hft_i)^2 
  =  \sum_{i \in \ulabqmoT  }  \ \sum_{  \xjt \sim \xit, \ j \in \ulabqT   }   (\ft_i - \hft_i)^2 \\
 & \leq \sum_{i \in \ulabqmoT  } \Kmaxqmo  \, (\ft_i - \hft_i)^2 
 = \Kmaxqmo  \ \|  \ft_{U_{\q-1}} - \hft_{U_{\q-1}} \|^2 .
\end{split}
\end{equation}
This gives in \eqref{eq:aq_haq_eq_terms}
\begin{equation*}
\sum_{j \in \ulabqT  } ( \af_j^{\q-1} +  \haf_j^{\q-1} + e_j^{\q-1} )^2  
\leq  \left (     
\sqrt{  \frac{\B_\q}{ \wmin_\q} }  + \sqrt{ \frac{\hB_\q}{ \wmin_\q} } 
+ \sqrt{ \Kmaxqmo }  \ \|  \ft_{U_{\q-1}} - \hft_{U_{\q-1}} \|
\right )^2.
\end{equation*}
Next, we bound the second term in \eqref{eq:fuq_error_v1} as
\begin{equation*}
\sum_{j \in \ulabqT  } \frac{1}{| \Neig_j^{\q-1}|}  
\leq   \sum_{j \in \ulabqT  }  \frac{1}{\Kminq}
= \frac{| \ulabqT |}{\Kminq}
\end{equation*}
where the inequality follows from the definition of $\Kminq$. Combining this with the bound on the first term in \eqref{eq:fuq_error_v1}, we get
\begin{equation*}
\|  \hft_{U_\q} - \ft_{U_\q}  \|^2 
\leq
 \frac{| \ulabqT |}{\Kminq}
 \left (     
\sqrt{  \frac{\B_\q}{ \wmin_\q} }  + \sqrt{ \frac{\hB_\q}{ \wmin_\q} } 
+ \sqrt{ \Kmaxqmo }  \ \|  \hft_{U_{\q-1}} - \ft_{U_{\q-1}} \|
\right )^2
\end{equation*}
which proves the lemma.
\end{proof}

\subsection{Proof of Lemma \ref{lem:targ_est_err}}
\label{pf:lem_targ_est_err}

\begin{proof}
The proof of Lemma \ref{lem:targ_est_err} is based on using the recursive relation provided in Lemma \ref{lem:targ_err_ito_prev} between the errors of consecutive layers. For brevity of notation, let us define
\begin{equation}
\begin{split}
c_q &=  \frac{| \ulabqT |}{\Kminq} \\
b_q &= \sqrt{  \frac{\B_\q}{ \wmin_\q} }  + \sqrt{ \frac{\hB_\q}{ \wmin_\q} } .
\end{split}
\end{equation}
Then, Lemma \ref{lem:targ_err_ito_prev} states that for $\q=1, \dots, \Q$,
\[
\|  \hft_{U_\q} - \ft_{U_\q}  \|  \leq   
\sqrt{ c_q } 
 \left (  
b_q
+ \sqrt{ \Kmaxqmo }  \ \|  \hft_{U_{\q-1}} - \ft_{U_{\q-1}} \|
\right ).
\]
Observing that the error of layer $\q=0$, which consists of the labeled nodes, is $0$ due to the assumption $ \hft_i = \ft_i $ for $i \in \labT$, we have 
\[
\|   \hft_{U_0} - \ft_{U_0}  \| =0.
\]
This gives
\begin{equation}
\begin{split}
\|  \hft_{U_1} - \ft_{U_1}  \|  
    &\leq \sqrt{ c_1 }   b_1  \\
\|  \hft_{U_2} - \ft_{U_2}  \| 
    &\leq    \sqrt{ c_2 }  b_2    +  \sqrt{ c_2 } \sqrt{K_1^{\max} }  \sqrt{ c_1 }   b_1   \\
\|  \hft_{U_3} - \ft_{U_3}  \| 
     &\leq  \sqrt{ c_3 }  (  b_3 +  \sqrt{K_2^{\max} } (\sqrt{ c_2 }  b_2    +  \sqrt{ c_2 } \sqrt{K_1^{\max} }  \sqrt{ c_1 }   b_1  )   ) \\
     &=    \sqrt{ c_3 }  b_3 +    \sqrt{ c_3 } \sqrt{K_2^{\max} } \sqrt{ c_2 }  b_2    +  \sqrt{ c_3 } \sqrt{K_2^{\max} } \sqrt{ c_2 } \sqrt{K_1^{\max} }  \sqrt{ c_1 }   b_1     
\end{split}
\end{equation}
Generalizing this, we get
\begin{equation*}
\|  \hft_{U_\q} - \ft_{U_\q}  \|  
\leq  \sum_{l=1}^\q \mu_l \, b_l
\end{equation*}
where $\mu_q =    \sqrt{ c_q }$ and
\begin{equation*}
\mu_l =   \sqrt{ c_q }  \prod_{m=l}^{\q-1} \left(  \sqrt{K_m^{\max} } \sqrt{ c_m } \right)
\end{equation*}
for $l=1, \dots, \q-1$.

Let us define the vectors $\muvect= [\mu_1 \, \mu_2 \,  \dots \, \mu_q]^T$ and $\bvect= [b_1 \, b_2 \,  \dots \, b_q]^T$. Then we can bound the error of the $\q$-th layer via the Cauchy-Schwartz inequality as
\begin{equation}
\label{eq:hft_ft_atq}
\|  \hft_{U_\q} - \ft_{U_\q}  \|^2 
\leq |  \langle  \muvect, \bvect  \rangle |^2
\leq \| \muvect \|^2 \, \| \bvect \|^2.
\end{equation}
Noticing that 
\begin{equation*}
b_q = \sqrt{  \frac{\B_\q}{ \wmin_\q} }  + \sqrt{ \frac{\hB_\q}{ \wmin_\q} } 
\leq \frac{1}{ \sqrt{\wmin} }  (\sqrt{\B_\q} + \sqrt{\hB_\q} )
\end{equation*}
and defining the vectors 
\begin{equation}
\begin{split}
C_\q &=[\sqrt{ \B_1} \, \sqrt{\B_2} \,  \dots \sqrt{ \B_\q} ]^T \\
\hat C_\q &=[\sqrt{ \hB_1} \, \sqrt{\hB_2} \,  \dots \sqrt{ \hB_\q} ]^T 
\end{split}
\end{equation}
we can bound the norm of  $\bvect$ as
\begin{equation}
\begin{split}
\| \bvect  \|  
&  \leq  
\frac{1}{\sqrt{ \wmin } } 
\| C_\q + \hat C_\q  \|
\leq  \frac{1}{\sqrt{ \wmin } }  (  \| C_\q \|  + \| \hat C_\q  \| ) \\
& \leq  \frac{1}{\sqrt{ \wmin } }  
\left( \sqrt{ \B_1 + \B_2 + \dots + \B_\q }
+
\sqrt{ \hB_1 + \hB_2 + \dots + \hB_\q  } \right) \\
& \leq  \frac{1}{\sqrt{ \wmin } }  
\left(  \sqrt{ (\ft)^T L^t \ft}
+\sqrt{ (\hft)^T L^t \hft}   \right) \\
& \leq \frac{1}{\sqrt{ \wmin } }  
\left( \sqrt{ \B } + \sqrt{\hB} \right).
\end{split}
\end{equation}

Using this in \eqref{eq:hft_ft_atq}, the total target estimation error can be bounded as
\begin{equation}
\begin{split}
\| \hft -\ft \|^2 
= \sum_{\q=1}^\Q  \|  \hft_{U_\q} - \ft_{U_\q}  \|^2 
 \leq  \sum_{\q=1}^\Q   \| \muvect \|^2 \, \| \bvect \|^2 \\
 \leq   \frac{1}{\wmin  }  
\left( \sqrt{ \B } + \sqrt{\hB} \right)^2 \ 
\sum_{\q=1}^\Q    \| \muvect \|^2 .
\end{split}
\end{equation}
Replacing $ \| \muvect \|^2$  with its open expression as
\begin{equation}
\begin{split}
 \| \muvect \|^2 
 &= \sum_{l=1}^\q \mu_l^2 
 =  \mu_\q^2 + \sum_{l=1}^{\q-1}  \mu_l^2 \\
 &= c_\q 
 +  \sum_{l=1}^{\q-1}     c_q   \prod_{m=l}^{\q-1}  K_m^{\max}  c_m  
 =  c_\q  
 \left( 1
 +  \sum_{l=1}^{\q-1}       \prod_{m=l}^{\q-1}  K_m^{\max}  c_m 
 \right) \\
 &= \frac{| \ulabqT |}{\Kminq} 
 \left( 1
 +  \sum_{l=1}^{\q-1}       \prod_{m=l}^{\q-1}  K_m^{\max}  \frac{|  I_{U_m}^t   |}{K_m^{\min}} 
 \right) 
\end{split}
\end{equation}
we get the stated result.

\end{proof}

\end{document}

%% file: elksty.tex
\newcommand{\half}{\ensuremath{\frac{1}{2}}}
\newcommand{\ones}{\ensuremath{\overrightarrow{1}}}
\newcommand{\rU}{\ensuremath{ \overline{U}}}

\newcommand{\ralpha}{\ensuremath{ \overline{\alpha}}}

\newcommand{\rLambda}{\ensuremath{ {\Lambda}}}

\newcommand{\labS}{\ensuremath{I_L^s}}
\newcommand{\labT}{\ensuremath{I_L^t}}   
\newcommand{\indT}{\ensuremath{I^t}}  
\newcommand{\ulabT}{\ensuremath{I_U^t}}
\newcommand{\ulabzerT}{\ensuremath{I_{U_0}^t}}
\newcommand{\ulaboneT}{\ensuremath{I_{U_1}^t}}
\newcommand{\ulabqT}{\ensuremath{I_{U_q}^t}}
\newcommand{\ulabqmoT}{\ensuremath{I_{U_{q-1}}^t}}
\newcommand{\ulabQT}{\ensuremath{I_{U_Q}^t}}
\newcommand{\q}{\ensuremath{q}}
\newcommand{\Q}{\ensuremath{Q}}
\newcommand{\sizeU}{\ensuremath{R}}
\newcommand{\R}{\ensuremath{\mathbb{R}}}
\newcommand{\N}{\ensuremath{N}}
\newcommand{\Ns}{\ensuremath{N_s}}
\newcommand{\Nt}{\ensuremath{N_t}}

\newcommand{\dlambda}{\ensuremath{\delta}}
\newcommand{\dalpha}{\ensuremath{\Delta_\alpha}}

\newcommand{\calpha}{\ensuremath{C}}
\newcommand{\lambdaR}{\ensuremath{\lambda_R}}
\newcommand{\hfs}{\ensuremath{\hat{f}^s}}
\newcommand{\hft}{\ensuremath{\hat{f}^t}}
\newcommand{\hhs}{\ensuremath{\hat{h}^s}}
\newcommand{\hht}{\ensuremath{\hat{h}^t}}
\newcommand{\fs}{\ensuremath{f^s}}
\newcommand{\ft}{\ensuremath{f^t}}
\newcommand{\Ms}{\ensuremath{\mathcal{M}^s}}
\newcommand{\Mt}{\ensuremath{\mathcal{M}^t}}
\newcommand{\Hs}{\ensuremath{\mathcal{H}^s}}
\newcommand{\Ht}{\ensuremath{\mathcal{H}^t}}
\newcommand{\gs}{\ensuremath{g^s}}
\newcommand{\gt}{\ensuremath{g^t}}
\newcommand{\lips}{\ensuremath{M_s}}
\newcommand{\lipt}{\ensuremath{M_t}}
\newcommand{\Al}{\ensuremath{A_l}}
\newcommand{\Au}{\ensuremath{A_u}}
\newcommand{\A}{\ensuremath{A}}
\newcommand{\xis}{\ensuremath{x_i^s}}
\newcommand{\xjs}{\ensuremath{x_j^s}}
\newcommand{\xit}{\ensuremath{x_i^t}}
\newcommand{\xjt}{\ensuremath{x_j^t}}
\newcommand{\Kis}{\ensuremath{K_i^s}}
\newcommand{\Kit}{\ensuremath{K_i^t}}

\newcommand{\Lphi}{\ensuremath{L_\phi}}
\newcommand{\dmin}{\ensuremath{d_{\min} }}
\newcommand{\dmax}{\ensuremath{d_{\max} }}
\newcommand{\emax}{\ensuremath{\epsilon_{\mathcal{N}}}}
\newcommand{\egen}{\ensuremath{\epsilon_{\Gamma}}}
\newcommand{\DelW}{\ensuremath{\Delta_W}}
\newcommand{\hB}{\ensuremath{\hat{B}}}
\newcommand{\B}{\ensuremath{B}}
\newcommand{\Kminq}{\ensuremath{ {K_q^{\min}} }}
\newcommand{\Kmaxq}{\ensuremath{ {K_q^{\max}} }}
\newcommand{\Kmaxqmo}{\ensuremath{ {K_{q-1}^{\max}} }}
\newcommand{\Neig}{\ensuremath{ \mathcal{N} }}
\newcommand{\wmin}{\ensuremath{ w^{\min} }}
\newcommand{\af}{\ensuremath{ a }}
\newcommand{\haf}{\ensuremath{ \hat{a} }}
\newcommand{\Af}{\ensuremath{ A }}
\newcommand{\hAf}{\ensuremath{ \hat{A} }}
\newcommand{\muvect}{\ensuremath{ \overrightarrow{\mu_q} }}
\newcommand{\bvect}{\ensuremath{ \overrightarrow{b_q} }}
\newcommand{\poly}{\ensuremath{ \mathrm{poly} }}

\newtheorem{theorem}{Theorem}

\theoremstyle{corollary}

\theoremstyle{lemma}
\newtheorem{lemma}{Lemma}

\theoremstyle{proposition}

\theoremstyle{axiom}

\theoremstyle{conjecture}

\theoremstyle{example}

\theoremstyle{definition}

\theoremstyle{remark}
